\documentclass{article}[11pt]
\usepackage{longtable}
\usepackage{booktabs}
\usepackage{latexsym,amsmath,amsfonts,amssymb} 
\usepackage{url}
\usepackage{graphicx} 
\usepackage{epstopdf}
\usepackage{algorithm,algorithmic}

\usepackage{fullpage}
\usepackage{hyperref}
\usepackage{natbib}
\usepackage{stmaryrd}
\usepackage{subcaption}
\usepackage{nicefrac}
\usepackage[absolute,overlay]{textpos}

\newcommand{\bigO}{\mathcal{O}}

\newcommand{\Algo}[1]{\textsc{#1}}

\newcommand{\figureBetweenHorizontal}{0pt}

\newcommand{\prob}{\mathbb{P}}

\newcommand{\argmax}{\operatornamewithlimits{argmax}}

\newcommand\R{\mathbb{R}}   

\newcommand{\given}{ \, \vert \, }
\renewcommand{\vec}[1]{\boldsymbol{#1}}

\newcommand{\cA}{{\cal A}} 
\newcommand{\cL}{{\cal L}}

\newcommand{\matS}{\mathbf{S}}
\newcommand{\matV}{\mathbf{V}}

\newcommand{\br}{\boldsymbol{r}}

\newcommand{\bv}{\boldsymbol{v}}

\newcommand{\bx}{\boldsymbol{x}}

\newcommand{\bX}{{\bf X}}

\newtheorem{Thm}{Theorem}

\newtheorem{Lem}[Thm]{Lemma}

\title{Online Preselection with Context Information\\ under the Plackett-Luce Model}
\author{ 
	\textbf{Adil El Mesaoudi-Paul, Viktor Bengs,  Eyke H\"ullermeier} \\
	 Heinz Nixdorf Institute and Department of Computer Science \\ 
	 Paderborn University, Germany\\
 	\{adil.paul,viktor.bengs,eyke\}@upb.de}
 
 \date{}
 
\begin{document}

\maketitle

\begin{abstract}
\noindent We consider an extension of the contextual multi-armed bandit problem, in which, instead of selecting a single alternative (arm), a learner is supposed to make a preselection in the form of a subset of alternatives. More specifically, in each iteration, the learner is presented a set of arms and a context, both described in terms of feature vectors. The task of the learner is to preselect $k$ of these arms, among which a final choice is made in a second step. In our setup, we assume that each arm has a latent (context-dependent) utility, and that feedback on a preselection is produced according to a Plackett-Luce model. We propose the \Algo{CPPL} algorithm, which is inspired by the well-known UCB algorithm, and evaluate this algorithm on synthetic and real data. In particular, we consider an online algorithm selection scenario, which served as a main motivation of our problem setting.  Here, an instance (which defines the context) from a certain problem class (such as SAT) can be solved by different algorithms (the arms), but only $k$ of these algorithms can actually be run.  
\end{abstract}

\section{Introduction} \label{sec:introdution}

In machine learning, the notion of \emph{multi-armed bandits} (MAB) refers to a class of online learning problems, in which a learner is supposed to simultaneously explore and exploit a given set of choice alternatives (metaphorically referred to as ``arms'') in the course of a sequential decision process \citep{LaSz18}. In this paper, we consider an extension of the basic setting, which is practically motivated by the problem of \emph{preselection} as recently introduced by \cite{SaGo18a} and \citet{bengs19}: Instead of selecting a single arm, the learner is only supposed to preselect a promising subset of arms. The final choice is then made by a \emph{selector}, for example a human user or another algorithm. In information retrieval, for instance, the role of the learner is played by a search engine, and the selector is the user who seeks a certain information. Another application, which served as a concrete motivation of our setting and will also be used in our experimental study, is the problem of algorithm (pre-)selection \citep{Kerschke18}. Here, the (presumably) best-performing algorithm needs to be chosen from a pool of candidates.
Our setting is related to, and partly builds on, various other extensions of the MAB problem that have been considered in the literature:
\begin{itemize}
	\item In the \emph{contextual setting}, the learner is provided with additional information about the context in which a choice is made \citep{Au02,La07,LiChLaSc10,ChLiReSc11}. The context may change from one iteration to the next, and has an influence on the usefulness (utility) of the choice alternatives (arms). For example, in the case of algorithm selection, each algorithm corresponds to an arm, and the context is specified by an instance of a certain problem class, e.g., a logical formula in SAT. 
	Obviously, the performance of an algorithm (in comparison to other algorithms solving the same problem) may strongly vary from instance to instance. 
	
	\item  
	In \emph{dueling} \citep{YuJo09} or \emph{preference-based} bandits \citep{BuHuEl18}, the assumption of numerical feedback is relaxed. Instead of selecting a single arm and observing a numerical reward, the learner selects a pair of arms and observes a qualitative comparison between these arms. This setting can be further generalized toward multi-wise comparisons, i.e., comparison of subsets of more than two arms, as recently have been studied under the notions of battling bandits \citep{SaGo18a} and preselection bandits \citep{bengs19}. Returning to our example of algorithm selection, qualitative feedback naturally occurs when running several algorithms on the same problem instance in parallel (e.g., on different cores of a CPU), and stopping the execution as soon as one of them returns a solution. 
	
\end{itemize}

\noindent In this paper, we address the problem of online preselection with context information, which can be seen as a contextual extension of the preselection problem as introduced by \citet{bengs19}. 

Starting with a more detailed overview of related work in the next section, we recall some important definitions and theoretical notions in Section~3. A formal description of the online preselection problem is then given in Section~
4, and an algorithm for tackling this problem is proposed in Section~5. An experimental study with both synthetic and real data is presented in Section~6, prior to   
concluding the paper with a summary and an outlook on future work in Section~7.

\section{Related Work}

The contextual bandit problem in the standard setting was first considered  by~\cite{Au02} under the notion of associative reinforcement learning with linear value functions. The author models the expected reward of an action in terms of the inner product of a feature vector (describing properties of the action in a certain context) and a weight vector, which is unknown but assumed to be the same for all actions. He proposes the \Algo{LinRel} algorithm, which achieves a regret bound of order $\tilde{\bigO} ( T^{1/2} )$.

The \Algo{Epoch-Greedy} algorithm (\cite{La07}) is an extension of the well-known $\epsilon $-greedy algorithm for the standard context-free bandit problem, where exploration and exploitation rounds are running in epochs. It achieves a regret of ${\bigO} ( T^{2/3} S^{1/3})$, where $S$ is the complexity term in a sample complexity bound for standard supervised learning. Other works on contextual MAB include \citep{LuPaPa10,AgHsKaLaLiSc14,AbPaSz11,AgGo13}.  

 \cite{LiChLaSc10} consider the linear contextual bandit problem with disjoint models, i.e., the weight vectors are specific to every action, and present the \Algo{LinUCB} algorithm, a UCB-like method for contextual bandit algorithms. An extended version of the algorithm for hybrid models, where some weight vectors are shared among the actions, is also presented. In their analysis of the regret of the algorithm, \cite{ChLiReSc11} show a high-probability regret bound of ${\bigO} \left( (T d \ln^3 (K T \ln(T) / \delta))^{1/2} \right)$. 
 
The notion of dueling bandits was introduced by \cite{YuJo09} and has been developed over the last decade, a recent survey of the field is provided by \cite{BuHuEl18}. 
In \cite{CoCr14} an online multi-task learning with a shared annotator is considered and for the proposed method it is shown that it can be used to solve the contextual dueling bandit problem.
Later on, \cite{DuHoShSlZo15} studied a scenario in which the learner iteratively observes a context, chooses a pair of arms, and observes the outcome of their comparison. The authors consider the solution concept of a von Neumann winner and present different algorithms for learning and  approximating such a winner, while minimizing regret. 

Closely related to the dueling bandit problem, and also relevant for our work, is learning from multi-wise comparisons. Different scenarios have been considered in the literature so far. \cite{SaGo18a} consider the regret minimization criteria under different subset choice models. In their setting, termed battling bandits, the learner selects in each round a multi-set of $k\ge 2$ arms and observes the winner of the selected subset. Its goal is to identify a best item and play it as often as possible. In a subsequent work, \cite{SaGo19a} study the sample complexity of the battling bandit problem, both for the case of winner feedback as well as feedback in the form of a full ranking over the selected subset. 
\cite{SaGo18b} examine the ranking problem under the PL model in the PAC setting. Here, the learner selects subsets of exactly $k$ items, and two feedback models are considered, the winner feedback as well as feedback in the form of an ordered list of $m\leq k$ items.
Considering winner feedback under the PL model, \cite{ReLiSh18} investigate the PAC top-$k$ and total ranking problems from $k$-wise ($k\ge 2$) comparisons, while \cite{ChLiMa18} study the sample complexity of identifying, with high probability, the top-$k$ ranking. 
\cite{SaGo19b} consider the regret minimization problem under the MNL model. They investigate two different settings. In the first, the learner selects subsets of size at most $k$ and obtains top-$k$ rank-ordered feedback. In the second, the learner selects subsets of a fixed size $k$ and observes a full ranking over the selected subset as feedback. 

A problem setup similar to ours is that of dynamic assortment optimization \citep{caro2007dynamic}, which has been studied in operational research. Here, the goal is to determine, from a set of products, the subset that maximizes the revenue, in the course of a dynamic interaction with different customers. Context information is taken into account by \citep{ChWaZh18,ChSi17,OuLiZhJi18}. As usual, the authors assign a real revenue to each product and assume an MNL choice model. Furthermore, they assume a linear relationship between product features and their utilities. 
%

Also similar to our setting is the problem of dyad ranking \citep{ScHu18}, which can be seen as a unification of different types of ranking problems that have been studied in the realm of preference learning, most notably so-called label and object ranking. The notion of a ``dyad'' refers to a combination of a context and a choice alternative, both of which are characterized in terms of features. \cite{ScHu18} tackle the problem of learning a dyad ranking function, that is, a function that preferentially sorts a set of candidate dyads based on their (joint) feature representation. To this end, they assume the Plackett-Luce model as an underlying model of probabilistic choice and ranking. In contrast to our work, however, they only consider the batch but not the online mode of learning.  

\section{Background and Notation}

For $n \in \mathbb N$, we denote by $[n]$ the set $\{1,\ldots,n\}$. We write $\| x \|$ for the Euclidean norm of a vector $x$, and $|| A ||_{\text{op}}$ for the operator norm, i.e., the largest singular value of a matrix $A$. For symmetric matrices $A,B$, we write $A \leq B$  if $B-A$ is positive semi-definite. Furthermore, we define the notions of ranking and ordering over a finite set of $n \in \mathbb N$ alternatives as well as partial rankings over subsets as follows:
\begin{itemize}
	\item We call a bijective mapping $\br:[n] \rightarrow [n]$ a \emph{ranking}, with the interpretation that $\br(k)$ is the rank of the $k^{\text{th}}$ alternative.
	The inverse function $\br^{-1}$, with $\br^{-1}(k)$ the index of the alternative on position $k$, is called the \emph{ordering induced by $\br$.} 
	\item A bijection $\br_S:S \rightarrow [\, |S|\, ]$ is called a full resp.\ partial ranking on $S\subseteq[n].$
	The inverse $\br_S^{-1}$ is called the ordering induced by $\br$ on $S$. For a partial ranking on $S \subseteq [n]$, we will drop the index if $S$  is clear from the context. 
	%
	%
\end{itemize}

\subsection{The Plackett-Luce Model}
The Plackett-Luce (PL) model is a parameterized probability model on the set of all rankings over a finite set of $n$ choice alternatives.
Its parameter $\bv=(v_1,\ldots,v_n)^{\top}\in \mathbb R^n_+$ represents the weights or latent utilities of the alternatives.
The probability mass function of the PL model is given by
$$	\prob(\br \given \bv) 
= \prod_{i=1}^{n} \frac{v_{\br^{-1}(i)}}{\sum_{j=i}^{n}
	v_{\br^{-1} (j)}} \, ,	
$$
where $\br$ is a ranking.
Note that the mode of the probability mass function is obtained for the ordering that sorts the parameter in descending order. 
Another appealing property of the PL model is that the probability of a marginal, i.e., a partial ranking $\br_S$ on a subset $S\subset [n]$, can be expressed in closed form:
\begin{align} \label{def:prob_partial_ranking}
\prob(\br_S \given  \bv)  = \!\!
\sum_{\br \in E(\br_S)} \! \!\!\! \prob(\br \given \bv) 
= \prod_{i=1}^{|S|} \frac{v_{\br_S^{-1}(i)}}{\sum_{j=i}^{|S|}
	v_{\br_S^{-1} (j)}} \, ,
%
\end{align}
with $E(\br_S)$ the set of linear extensions of $\br_S$ and $\br_S^{-1}$ the ordering induced by $\br_S$. Likewise, the probability that alternative  $k\in S$ gets the top rank is
\begin{align} \label{def:prob_top_rank}
\prob(\br_S(k)=1 \given  \bv) 
=  \frac{v_{k}}{\sum_{i\in S}
	v_{i}} \,.
\end{align}

\subsection{Plackett-Luce Model with Feature and Context Information}

To incorporate feature information $\vec{x}_i \in \mathbb{R}^d$ about the $i^{th}$ choice alternative, we follow the approach by \citep{cheng2010label,ScHu18} and replace the constant latent utility $v_i$ by a log-linear function 
\begin{align} \label{def:utility_param}
v_i = v_i(\vec{x}_i) = \exp \left( \theta^{\top} \vec{x}_{i} \right)  \,  .			
\end{align}
Given $n$ choice alternatives, which define a contextual decision problem, we summarize the corresponding feature vectors $\vec{x}_1, \ldots , \vec{x}_n$ in a matrix $\bX \in \mathbb{R}^{d \times n}$ and write
$$	v_i = v_i(\bX) = \exp \left( \theta^{\top} \vec{x}_{i} \right), \quad i \in \{1,\ldots,n\} .		$$ 
With this, we define the PL model with context information by
\begin{align*}		
\prob (\br \given \theta, \bX) 
& = \prod_{i=1}^{n} \frac{v_{\br^{-1}(i)}(\bX)}{v_{\br^{-1}(i)}(\bX)+ \cdots + v_{\br^{-1}(n)}(\bX)} 
  = \prod_{i=1}^{n} \frac{ \exp \left( \theta^{\top} \bx_{\br^{-1}(j)} \right) }{ \sum_{j=i}^n \exp \left( \theta^{\top} \bx_{\br^{-1}(j)} \right) }.
\end{align*}
The marginals, i.e., the probability of a partial ranking defined in (\ref{def:prob_partial_ranking}), on a subset $S\subset [n]$, will be denoted by 
\begin{align} \label{def:marginals_context}
P_\theta(\sigma \given  S, \bX)  
= \prod_{i=1}^{|S|} \frac{\exp \left( \theta^{\top} \bx_{\sigma^{-1}(i)} \right)}{\sum_{j=i}^{|S|} \exp \left( \theta^{\top} \bx_{\sigma^{-1}(j)} \right)} \enspace ,
\end{align}
where $\sigma$ is a partial ranking on $S$ and $\sigma^{-1}$ is the ordering induced by $\sigma.$
For sake of brevity, we will subsequently suppress the subset $S$ on which the partial ranking $\sigma$ is defined.
%

For an observation $(\sigma,S,\bX)$ the log-likelihood function  under this probability model is given by 
\begin{flalign} \label{def:log_likelihood}
\begin{split}
&\cL \left( \theta \given \sigma,S,\bX \right)  = \sum_{i=1}^{|S|} \left[ \theta^{\top} \bx_{\sigma^{-1}(i)} - \log \left( \sum_{j=i}^{|S|} \exp \left( \theta^{\top} \bx_{\sigma^{-1}(j)} \right) \right) \right] \enspace .
\end{split}
\end{flalign}  

%
%

For an alternative $k\in S$, the probability that it gets the top rank among the alternatives in $S,$ i.e., the top-rank probability in (\ref{def:prob_top_rank}), will be denoted by
\begin{align} \label{def:top_rank_context}
P_\theta(k \given  S, \bX)  
= \frac{\exp \left( \theta^{\top} \bx_{k} \right)}{\sum_{j \in S} \exp \left( \theta^{\top} \bx_{j} \right)} \enspace.
\end{align}
The corresponding log-likelihood function for an observation $(k,S,\bX)$ is 
\begin{align} \label{def:log_likelihood_top_rank}
\cL \left( \theta \given k ,S,\bX \right) 
&=  \theta^{\top} \bx_{k} - \log \left( \sum_{j \in S} \exp \left( \theta^{\top} \bx_{j} \right) \right)   \enspace,
\end{align}  
where $k \in S.$
The concavity of (\ref{def:log_likelihood}) resp.\ (\ref{def:log_likelihood_top_rank}) was shown by \citet{ScHu18}.

\section{Contextual Online Preselection}


In the following, we define the contextual online preselection problem in a formal way.
To this end, we consider a set of $n \in \mathbb N$ available choice alternatives that we refer to as \emph{arms}, and simply denote them by their index: $\cA = \left\{ 1, \dots , n \right\}$.
The learning problem proceeds in a possibly infinite time horizon $T,$ where in each time step $t\in \{1,\ldots,T\}$, the learner observes a  context  $\bX_t =(\bx_{t,1} \dots \, \bx_{t,n})$ with $\bx_{t,i} \in \R^{d}$ for  any arm $i$. 
Each vector $\bx_{t,i}$ encodes features of the context in which an arm must be chosen, but possibly also of the arm $i$ itself. In other words, $\bx_{t,i}$ contains properties of both the context and the arm, for instance obtained by a joint feature map.
%
After observing the context information, the learner selects a subset $S_t \subseteq [n]$ of $k<n$ distinct arms, and obtains feedback in the form of either one of the following:
\begin{itemize}
	\item A ranking over these arms, which is assumed to be generated by (\ref{def:marginals_context}), that is, a marginal of a PL model with context information $\bX_t = (\bx_{t,1} \dots  \bx_{t,n})$ and some \emph{unknown} weight parameter $\theta^* \in \mathbb R^d$ (partial ranking feedback scenario). 
	\item The top-ranked arm among these arms, which is assumed to be generated by (\ref{def:top_rank_context}), that is, a top-arm marginal of a PL model with context information $\bX_t = (\bx_{t,1} \dots  \bx_{t,n})$ and some \emph{unknown} weight parameter $\theta^* \in \mathbb R^d$ (winner feedback scenario). 
\end{itemize}
%
%
%
The goal of the learner is to select, in each time step $t$, a subset $S_t$ of arms that contains the arm which is best for the current context $\bX_t.$
In the realm of preference-based multi-armed bandits, the notion of a best arm can be defined in various ways \citep{BuHuEl18}.
In our setting, where we assume choices to be guided by the Plackett-Luce model (with fixed but unknown weight parameter $\theta^*$), it is natural to define the best arm for a time step $t$ by the arm with the highest (latent) utility (\ref{def:utility_param}).
More specifically, the best arm for the current time step $t$ is 
\begin{align}
i^*(t) = \argmax_{i \in \cA } v_{i}^*(\bX_t) = \argmax_{i \in \cA } \, \exp \left(  \vec{x}_{i}^{\top} \theta^* \right).
\end{align}
%
This definition is also in accordance with other conceivable definitions of a best arm, for instance the Borda winner for a time-dependent resp.\ context-dependent Borda score.
%
%

The regret of a learner selecting subset $S_t$ at time $t$ is set to be
$$
r_t = \frac{  v_{i^*(t)}^*(\bX_t) - \max_{j \in S_t} v_{j}^*(\bX_t) }{	v_{i^*(t)}^*(\bX_t)} \, ,
$$
which means that the cumulative regret for selecting subsets $(S_t)_{t\in[T]}$ during the time horizon $T$ is given by
\begin{align}\label{regret_def} 
R_T \, = \, \sum_{t=1}^{T} r_t \, = \, T - \sum_{t=1}^{T} \frac{ \max_{j \in S_t} v_{j}^*(\bX_t) }{	v_{i^*(t)}^*(\bX_t)} \, . 
\end{align}
Again, this notion of regret appears to be natural in our setting. It penalizes the learner with the relative difference between the utility of the truly best arm $i^*(t)$ and the utility of the best arm included in $S_t$\,---\,or, equivalently, the absolute difference after normalizing the utility of $i^*(t)$ to 1 (which is necessary, because the PL model is only defined up to a multiplicative factor). 

Note that the regret vanishes if $i^*(t) \in S_t,$ so that this notion of regret can be interpreted as a contextual version of the weak regret considered in the dueling bandits problem \citep{chen2017dueling}, where no regret occurs as long as the best arm is involved in the duel.

\section{The CPPL Algorithm}

At the core of the learning task is the estimation of the unknown parameter $\theta^*$, which basically determines the underlying contextual PL model of the feedback mechanism for both variants (partial ranking and winner feedback).
Thanks to the parametric form of the contextual PL model, one obtains a natural loss in terms of the log-likelihood function (see (\ref{def:log_likelihood}) resp.\ (\ref{def:log_likelihood_top_rank})), so that sensible estimates for $\theta^*$ can be obtained by minimizing this loss function in each feedback scenario, respectively.
This minimization problem can be solved by using suitable optimization methods, such as variants of the stochastic gradient descent (SGD) algorithm as used in our method below.

In addition to the estimation of the unknown parameter, the nature of the online learning task calls for tackling the well-known exploration-exploitation problem.
To this end, we adopt an approach similar to that of the UCB algorithm by deriving confidence bounds on the contextualized utility parameters in (\ref{def:utility_param}), and selecting the subset with maximal upper bounds.
Before stating our algorithm at the end of this section, we describe some important theoretical properties that it builds on.

\subsection{Theoretical Properties of SGD}
For the optimization of the loss function, i.e., the maximization of the log-likelihood function, we intend to use the stochastic gradient descent method in the form of the Polyak-Ruppert averaged SGD \citep{Ru88,PoJu92}.
It is defined by 
\begin{align} \label{def:SGD_iid}
\bar \theta_t =  \frac1t \sum_{i=1}^{t} \hat \theta_i \, , \quad \mbox{with} \quad \hat \theta_i =  \hat \theta_{i-1} - \gamma_t \nabla \l(\hat \theta_{i-1}; Z_i) \, ,
\end{align}
where $\l(\cdot;\cdot)$ is some loss function, $Z_i$ denotes the single observation in iteration step $i$, and $\gamma_t$ is the learning rate.
Under certain assumptions on $\l,$ \cite{FaXuYa18} show that $\sqrt{t}(\bar \theta_t - \theta^*) $ is asymptotically normal with a covariance matrix $\Sigma$ of the form $ \matS \matV^{-1} \matS,$ where $\theta^* \in \mathbb R^d$ is the minimizer of $\mathbb E [\l(\theta;Z)]$ and $Z$ has the same distribution as the observations $Z_i.$ 
The components $\matS$ and $\matV$ of the covariance matrix depend on the gradient and the Hessian matrix of the loss function $\l,$ for which  \cite{FaXuYa18} propose plug-in estimates $\hat{\matS}$ and $\hat{\matV}$ in the case of a twice differentiable loss function $\l.$
Having this asymptotic normality, it is a well-known result that under certain regularity conditions on $\l$, the plug-in estimates $\hat{\matS}$ and $\hat{\matV}$ are consistent and the Wald-type statistic 
\begin{align} \label{def:wald_type_statistic}
d^{-1}(\bar \theta_t - \theta^*)^{\top} \, t \, \hat{\matS} \hat{\matV}^{-1} \hat{\matS} \, (\bar \theta_t - \theta^*)
\end{align}
converges in distribution to an F-distribution with degrees of freedom $d$ and $t-d,$ which for $t \rightarrow \infty$ in turn converges to a $\chi^2$-distribution with $d$ degrees of freedom.

\subsection{Confidence Sets for F-Distributions}
Given that the asymptotic distribution of the Wald-type statistic composed of the Polyak-Ruppert averaged SGD estimate is an F-distribution, a reasonable way to derive (asymptotic) confidence sets for the latter statistic is by analyzing confidence sets for the F-distribution.

For this purpose, we have the following lemma, which we prove in the supplementary material.
\begin{Lem}\label{eq:F_distr_tails} For a random variable $X$ with F-distribution with degrees of freedom $d_1$ and $d_2$ it holds that
	\begin{flalign}
	\begin{split}
	&\prob\left( X \geq \frac{ 4(d_1 + 2\sqrt{d_1\,x} + 2x)}{3\, d_1} \right) 
	\leq \exp(-x) + \exp\left( - \frac{3\, d_2}{2^8} \right), \quad x\geq 0 \enspace .
	\end{split}
	\end{flalign} 
\end{Lem}

\subsection{Confidence Sets for Contextualized Utility Parameters}
%
For our algorithm, we will use the Polyak-Ruppert averaged SGD method with the negative log-likelihood as a loss function.
Due to the contextualized online scenario and the active decision strategy of the learner, the losses observed at different time steps are not independent.
Indeed, note that at time step $t$ the observation consists of a triple $(Y_t,S_t,\bX_t),$ where $\bX_t$ is the context information, $S_t\subset [n]$ is the $k$-sized subset selected by the learner, and $Y_t \sim P_\theta(\cdot \given  S_t, \bX_t) $ with
\begin{itemize}
	\item $Y_t = \sigma_t,$ i.e., a partial ranking on the set $S_t$ provided by the underlying contextualized PL model (\ref{def:marginals_context}) in the case of partial ranking feedback, 
	\item $Y_t = k_t\in S_t,$ i.e., the top-ranked arm among the subset $S_t$ provided by the underlying contextualized PL model  (\ref{def:top_rank_context}) in the case of winner feedback.
\end{itemize}
However, since the learner determines its selection $S_t$ based on the entire history, i.e., the subsets selected so far, the observations will normally not be independent.
In addition, the contexts $\bX_t$ are not necessarily i.i.d.\ either.

Like similar approaches, we nevertheless use the asymptotic results for the i.i.d.\ case as an approximation. More specifically, for deriving confidence bounds on the contextualized utility parameters, we use the above results for the Polyak-Ruppert averaged SGD method in (\ref{def:SGD_iid}) as follows.
For some $\alpha \in(\nicefrac12,1)$ and $\gamma_1>0$, let
\begin{align*}
\bar{\theta}_t &= (t-1) \bar{\theta}_{t-1}/t + \hat{\theta}_{t}/t \, , \\
\hat{\theta}_{t} 
&= \hat{\theta}_{t-1} + \gamma_1t^{-\alpha} \nabla \cL \left( \hat{\theta}_{t-1} \vert Y_t, S_t, \bX_t \right), 
\end{align*}
be the Polyak-Ruppert averaged SGD estimate for $\theta^*,$ where the gradient of the log-likelihood function (as well as the Hessian) are given in the supplementary material.
Then, we estimate the true unknown contextualized utility parameter of an arm $i \in \cA$  at time step $t,$ i.e., $v_{i}^*(\bX_t) = \exp\left( \bx_{t,i}^{\top} \theta^* \right),$ by
\begin{align} \label{eq:estimate_utility_param}
\hat{v}_{t,i} = \exp \left(  \bx_{t,i}^{\top} \bar{\theta}_t \right).
\end{align}
For these estimates, we conclude that 
$	| \hat{v}_{t,i} - v_{i}^*(\bX_t) |  \leq c_{t,i}	$
holds with high probability for the i.i.d.\ case, where  $\omega>0$ is some suitable constant and 
\small
\begin{align*} 
\begin{split}
c_{t,i} &= \omega \,\sqrt{ \Big(  2 \log(t) + d + 2 \sqrt{d\, \log(t)} \,  \Big) \, \hat I_t } \\
\hat I_t &=  || \hat{\Sigma}_t^{1/2} M_t^{(i)}(\bar{\theta}_t) \hat{\Sigma}_t^{1/2} ||_{\text{op}}  \, , \\
M_t^{(i)}(\theta) &= \exp\left( 2 \bx_{t,i}^{\top} \theta \right) \bx_{t,i} \bx_{t,i}^{\top}, \\
\hat{\Sigma}_t &= t^{-1} \, \hat{\matS}_t^{-1} \,  \hat{\matV}_t \, \hat{\matS}_t^{-1} \\
\hat{\matS}_t &= \frac1t \sum_{i=1}^{t}  \nabla^2 ~\cL \left( \bar{\theta}_i \vert Y_i, S_i, \bX_i\right), \\
\hat{\matV}_t &= \frac1t \sum_{i=1}^{t}  \nabla ~\cL \left( \bar{\theta}_i \vert Y_i, S_i, \bX_i\right) \nabla ~\cL \left( \bar{\theta}_i \vert Y_i, S_i, \bX_i\right)^{\top}.
%
%
%
%
\end{split}
\end{align*}
\normalsize
This can be seen as follows. 
Define $f_{t,i}(\theta) = \exp\left( \bx_{t,i}^{\top} \theta \right)  $ and note that $v_{i}^*(\bX_t)=f_{t,i}(\theta^*)$ and $\hat{v}_{t,i}=f_{t,i}(\bar{\theta}_t) $. 
Thus, the mean value theorem implies the existence of $\tilde{\theta}_t = \bar{\theta}_t + \lambda(\bar{\theta}_t-\theta^*)$ for some $\lambda \in [0,1]$, such that
\begin{flalign}\label{eq_mvt}
\begin{split}
&| \hat{v}_{t,i} - v_{i}^*(\bX_t) | 
= | f_{t,i}(\bar{\theta}_t) - f_{t,i}(\theta^*) | 
= \sqrt{ (\bar{\theta}_t - \theta^* \, )^{\top} \left[ \nabla_{\theta} f_{t,i}(\tilde{\theta}_t) \nabla_{\theta} f_{t,i}(\tilde{\theta}_t)^{\top} \right] (\bar{\theta}_t - \theta^*) } \, , 
\end{split}
\end{flalign} 
where straightforward calculations show that
\begin{align*}
\nabla_{\theta} f_{t,i}(\tilde{\theta}_t) \nabla_{\theta} f_{t,i}(\tilde{\theta}_t)^{\top} = M_t^{(i)}(\tilde{\theta}_t) \, .
\end{align*}
Since the asymptotic normality of $\bar{\theta}_t$ implies its consistency for $\theta^*$, one can find a suitable constant $C>0$, such that, for $t$ sufficiently large,  
\begin{align*} 
%
|| M_t^{(i)}(\tilde{\theta}_t) - M_t^{(i)}(\bar{\theta}_t) ||_{\text{op}} \leq C
\end{align*}
holds with high probability.
Thus, for $t$ sufficiently large,  (\ref{eq_mvt}) can be bounded due to the asymptotic behavior of the Wald-type statistic (cf.\ (\ref{def:wald_type_statistic})) by the latter two displays and Lemma \ref{eq:F_distr_tails} as  
\begin{flalign*} 
&| \hat{v}_{t,i} - v_{i}^*(\bX_t)| 
\leq  \Big(  \sqrt{\hat I_t } +\sqrt{C}\Big) \sqrt{ (\bar{\theta}_t - \theta^*)^{\top}  \hat{\Sigma}_t^{-1} (\bar{\theta}_t - \theta^*) } \leq  c_{t,i}
\end{flalign*}
with high probability for some appropriate constant $\omega>0.$
Thus, for any arm $i \in \cA$, we conclude that $\hat{v}_{t,i} +  c_{t,i} $ is an asymptotic upper confidence bound for $v_{i}^*(\bX_t)$ for the i.i.d.\ case,  provided $t$ is sufficiently large.


\subsection{The \Algo{CPPL} Algorithm}

Our algorithm for the contextual preselection problem under the PL model (\Algo{CPPL}) is shown in Algorithm \ref{alg:mle_top}. 
As already explained, it adopts a strategy similar to that of UCB based on the confidence bounds for the contextualized utility parameters as derived above.

We start with a random initialization for the parameter vector.
In each time step $t=1,2,\ldots,T$, the context vectors are revealed, and the contextualized utility parameter of all arms are computed based on the current estimate of the parameter vector. 
Then, the $k$ arms with the highest upper bounds on the latent utility are selected, for which $Y_t$ is revealed according to the considered feedback scenario.
%
Finally, the estimate for the parameter is updated according to the Polyak-Ruppert averaged SGD method based on the gradient of the log-likelihood function for the observed triple, again depending on the considered feedback scenario.

Note that the update rule is computationally favorable, as it only involves some matrix operations which cause costs of order $O(d^3),$ since only the current observation $Y_t$ is processed for the optimization of the log-likelihood function.
This is particularly advantageous compared to optimizing the log-likelihood function based on the whole history in each time step, as for instance proposed in  \citet{ChWaZh18} in the related dynamic assortment optimization problem.

\begin{algorithm}
	\caption{\Algo{CPPL($n,k,T,\gamma_1,\alpha,\omega$)}}\label{alg:mle_top}
	\begin{algorithmic}[1]
		\STATE Initialize $\hat{\theta}_0$ randomly
		\STATE $\bar{\theta}_0=\hat{\theta}_0$ 
		\FOR{$t=1,2,\ldots,T$} 
		\STATE Observe the context vectors $\bX_t =(\bx_{t,1} \ldots \bx_{t,n})$
		\STATE Compute the estimated contextualized utility parameters $\hat{v}_{t,i}$  by means of (\ref{eq:estimate_utility_param})
		\STATE Choose $S_t$ as: 
		\begin{align*}
		\argmax_{S_t \subseteq [n],~|S_t|=k} \, \sum_{i\in S_t} \big(\hat{v}_{t,i} + c_{t,i} \big)
		\end{align*} 
		\STATE Observe $Y_t$ according to the feedback scenario
		\STATE Update $\bar{\theta}_t$ by 
		$\bar{\theta}_t = (t-1) \bar{\theta}_{t-1}/t + \hat{\theta}_{t}/t $ with
		\begin{align*}
		\hat{\theta}_{t} 
		&= \hat{\theta}_{t-1} + \gamma_1t^{-\alpha} \nabla \cL \left( \hat{\theta}_{t-1} \vert Y_t, S_t, \bX_t\right)
		\end{align*}
		\ENDFOR
	\end{algorithmic}
\end{algorithm}

\section{Experiments} \label{sec_exp}
In this section, we present experimental results for our learning algorithm on synthetic data as well as on an algorithm selection problem.

\subsection{Synthetic Data}\label{synthDataExp}
For the experiments on synthetic data, we consider various scenarios with respect to the number of available arms $n,$ context vector dimension $d,$  and size of the selection $k.$
For each scenario, the parameter resp.\ context vectors were sampled at random from the unit interval resp.\ unit square. 
%
%
%
%
%

For comparison, we included the \Algo{Max-Theta} method, which chooses at each time step the subset with maximal performance according to the current estimate of $\theta^*$, that is, line 6 in Algorithm \ref{alg:mle_top} is substituted by 
$\argmax_{S_t \subseteq [n],~|S_t|=k} \, \sum_{i\in S_t} \big(\hat{v}_{t,i} \big)$, the \Algo{$\epsilon$-Greedy} method, which adopts the choice of \Algo{Max-Theta} method with probability $1-\epsilon$ and a random choice otherwise ($\epsilon$ was set to $0.1$ in all experiments), and \Algo{MM}, a context-free PL-based method, in which the parameters of the PL model are estimated using the MM algorithm \citep{hunter2004mm}.  
Our \Algo{CPPL} method is instantiated with the parameters $\gamma_1 = 2, \alpha = 0.6, \omega =1.$

\begin{figure}
	\vspace{\figureBetweenHorizontal}
	\begin{subfigure}{.333\textwidth}
		\centering
		\includegraphics[width=.99\linewidth]{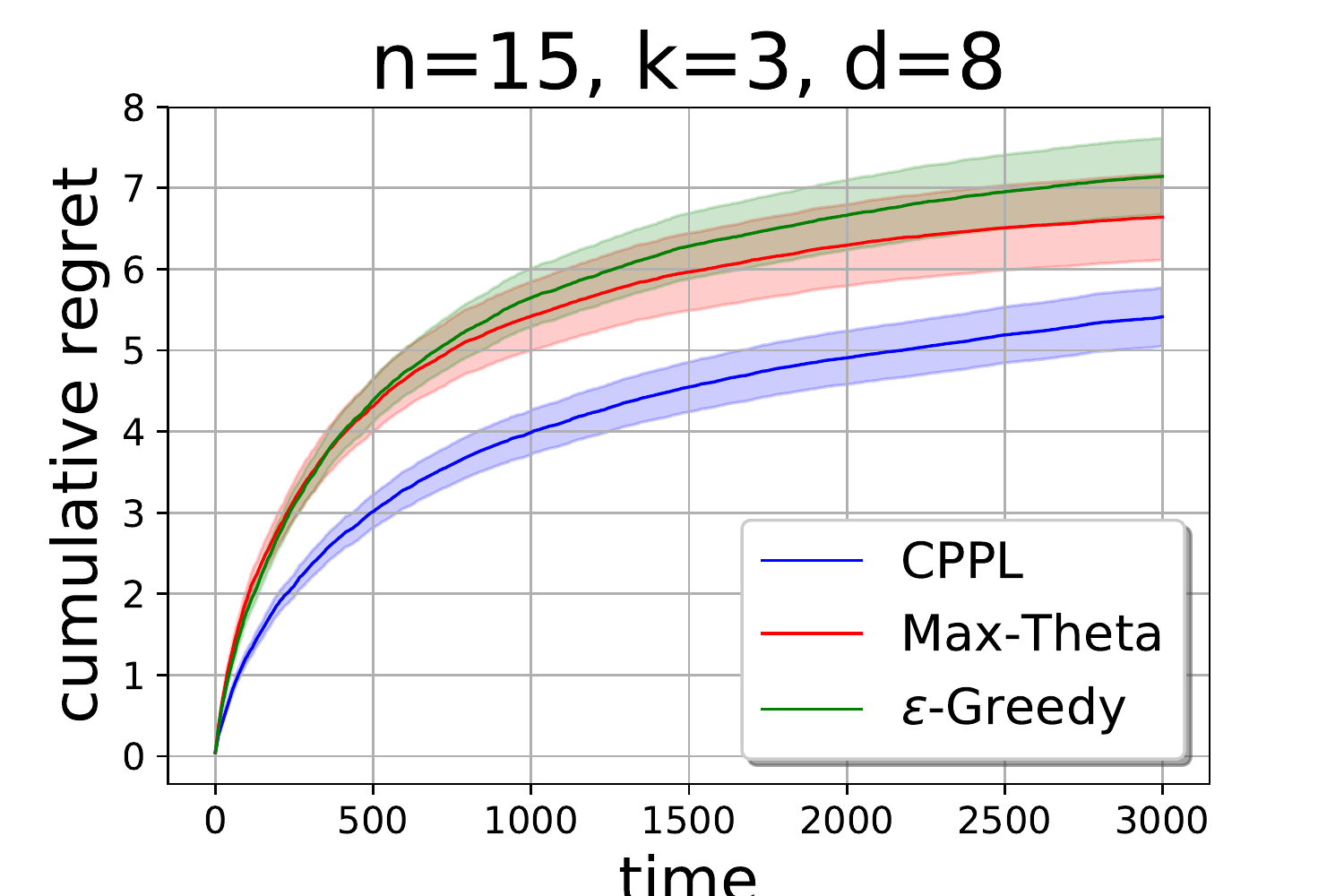}
	\end{subfigure}%
	\begin{subfigure}{.333\textwidth}
		\centering
		\includegraphics[width=.99\linewidth]{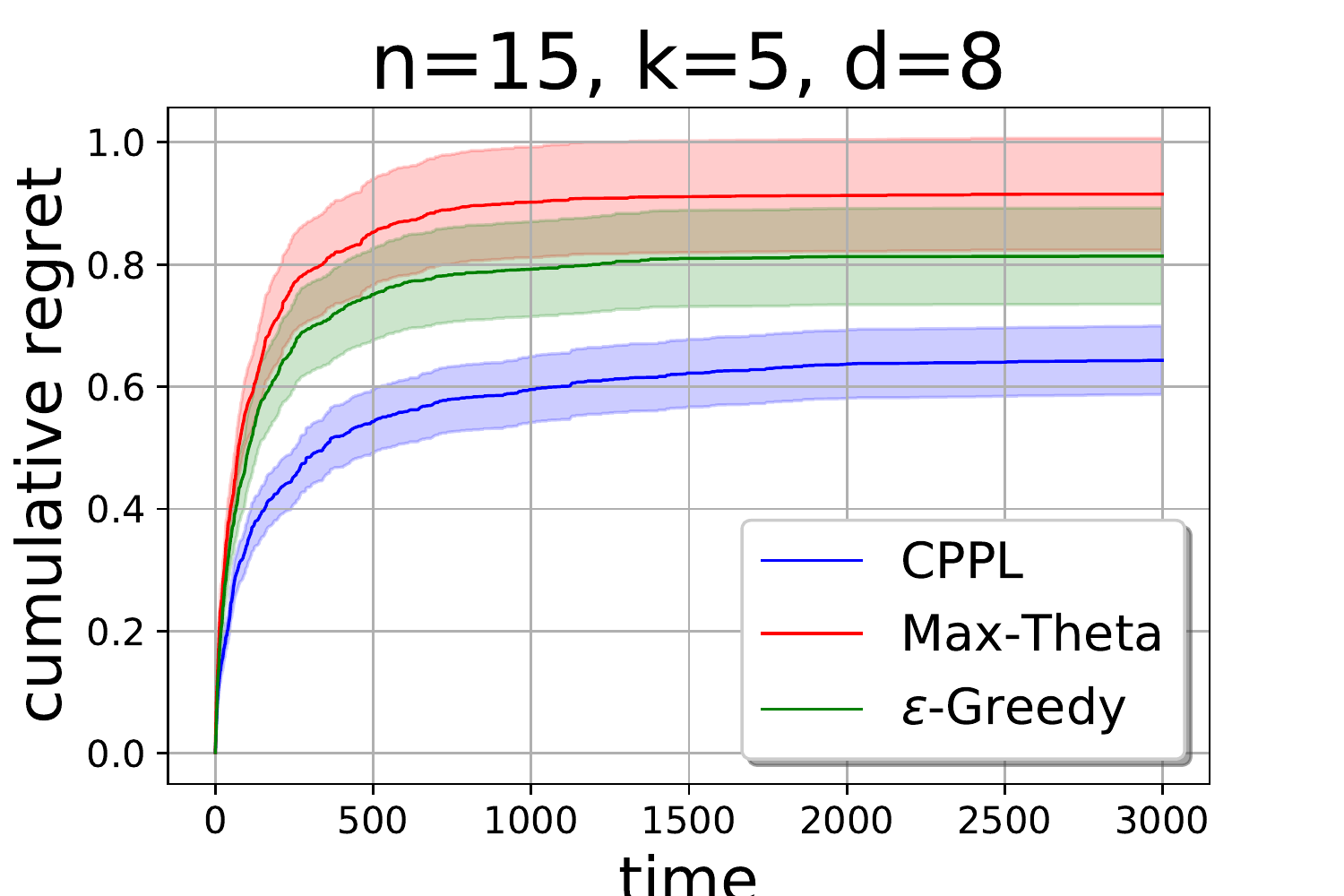}
	\end{subfigure}
	\begin{subfigure}{.333\textwidth}
		\centering
		\includegraphics[width=.99\linewidth]{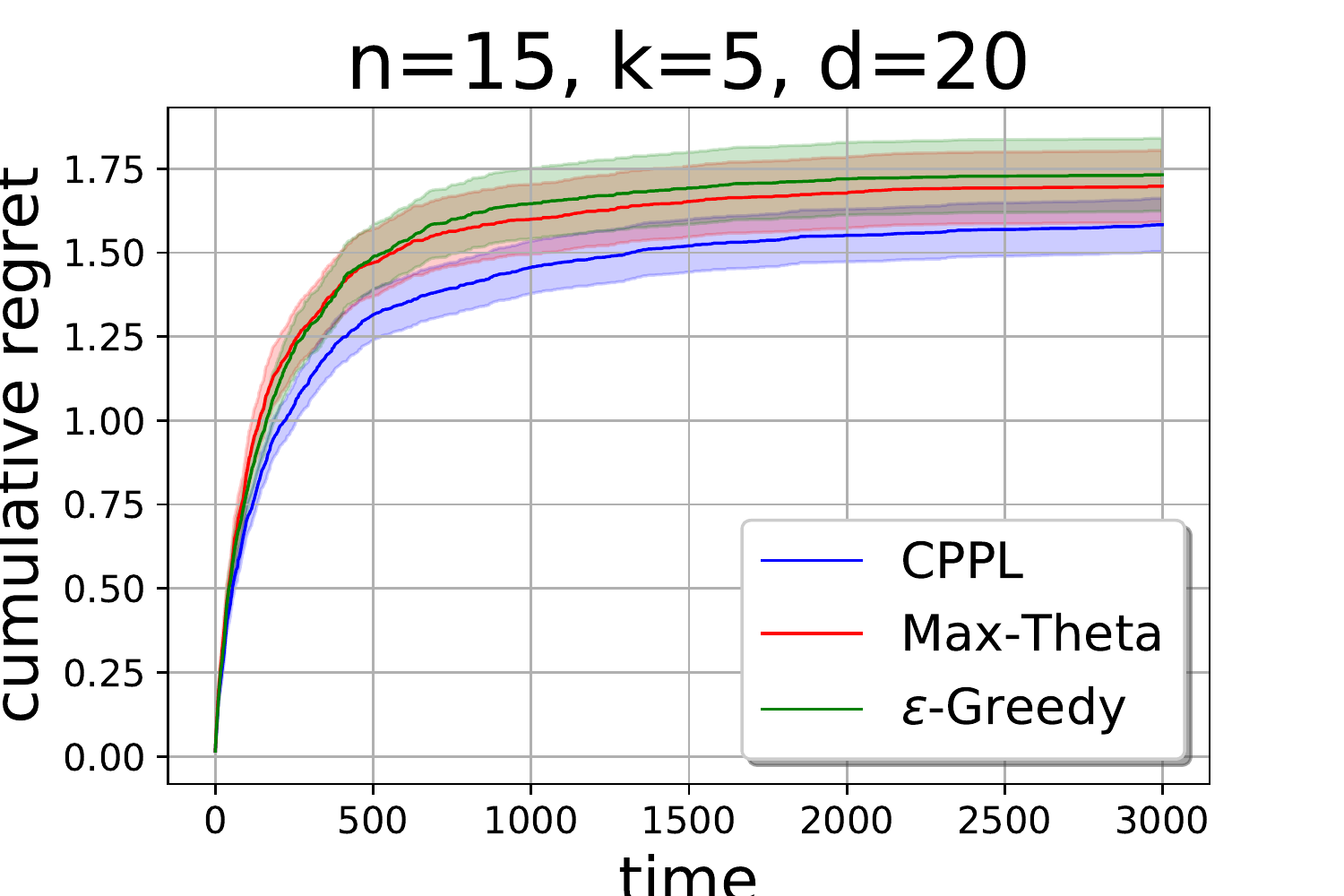}
	\end{subfigure}
	\begin{subfigure}{.333\textwidth}
		\centering
		\includegraphics[width=.99\linewidth]{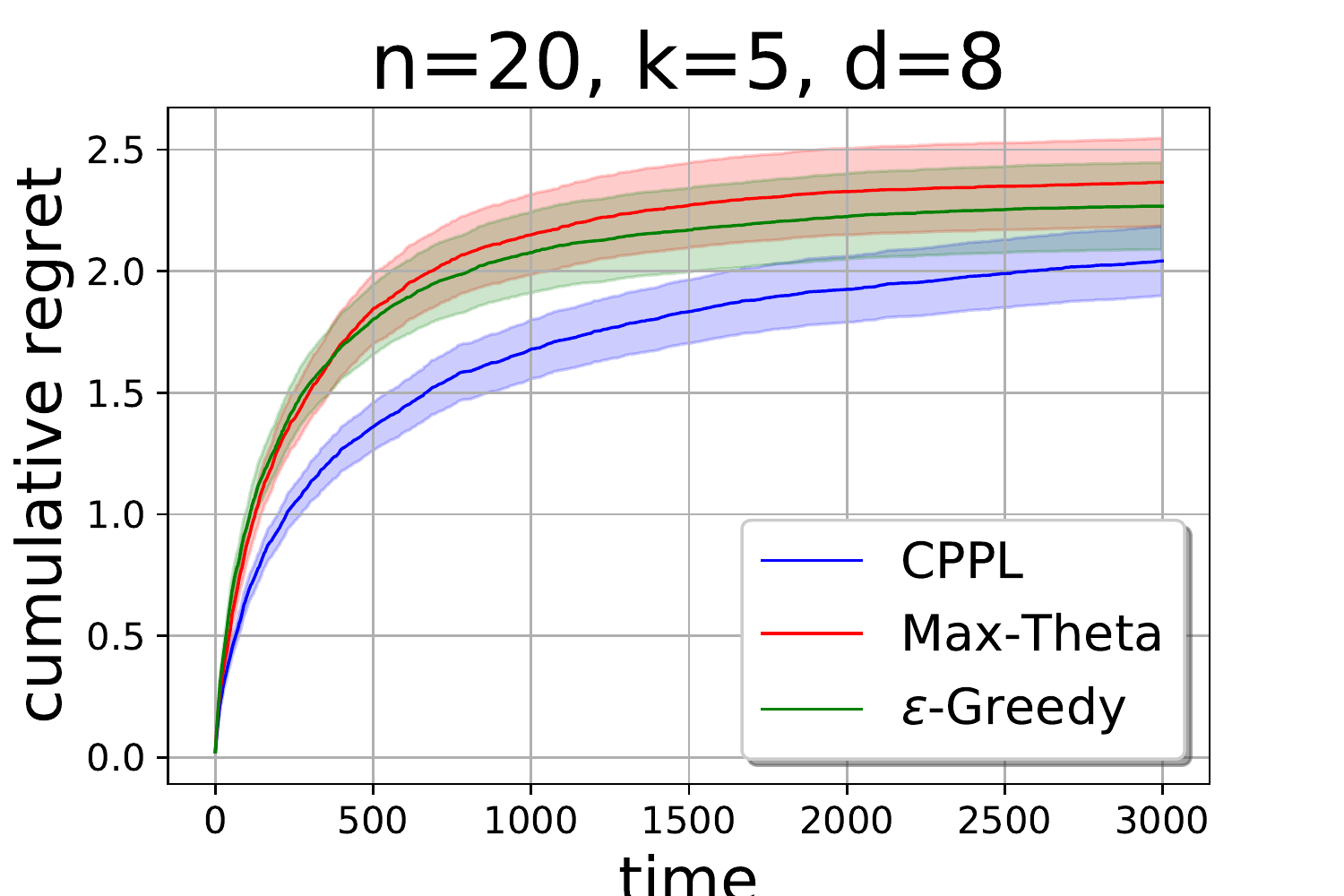}
	\end{subfigure}%
	\begin{subfigure}{.333\textwidth}
		\centering
		\includegraphics[width=.99\linewidth]{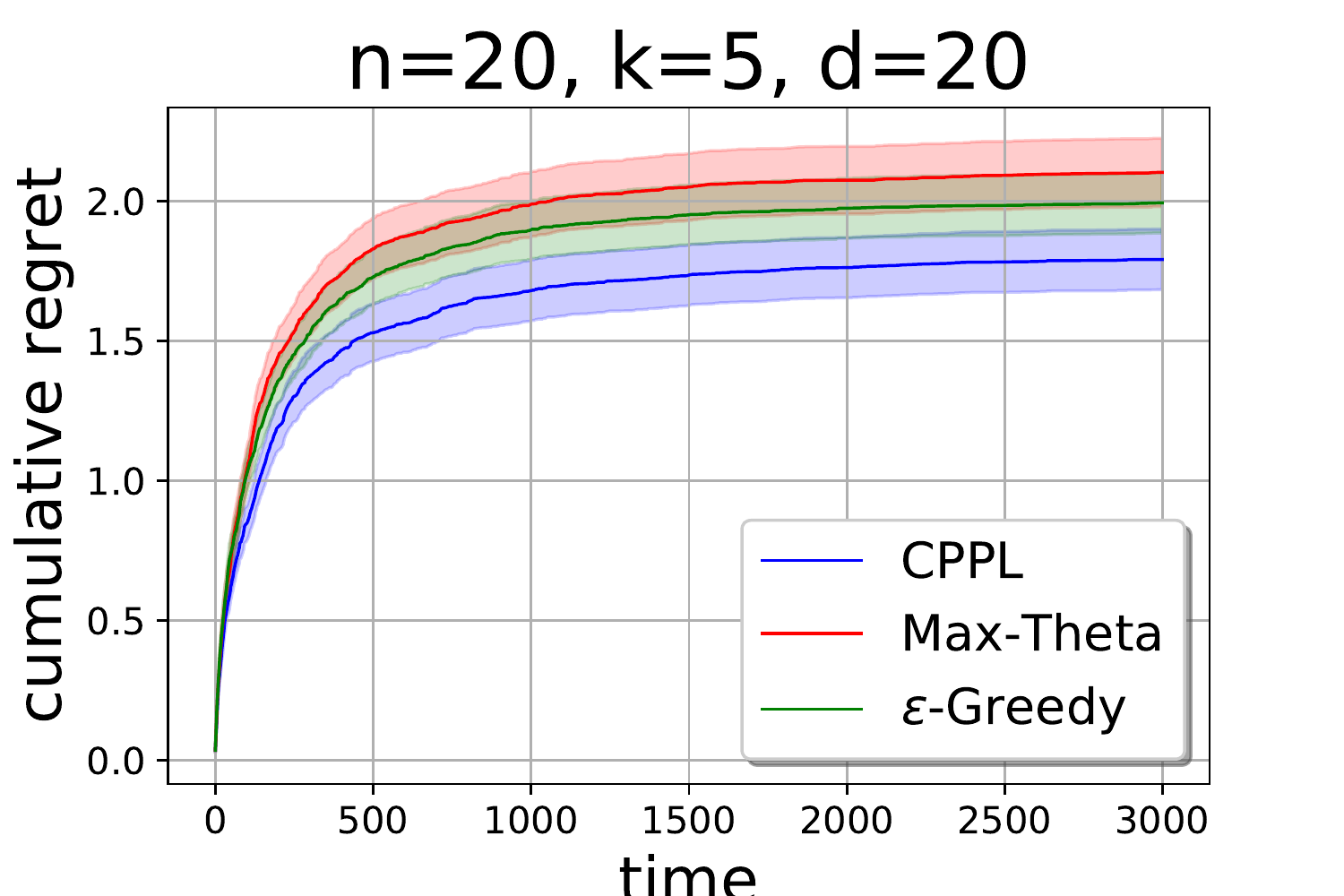}
	\end{subfigure}
	\begin{subfigure}{.333\textwidth}
		\centering
		\includegraphics[width=.99\linewidth]{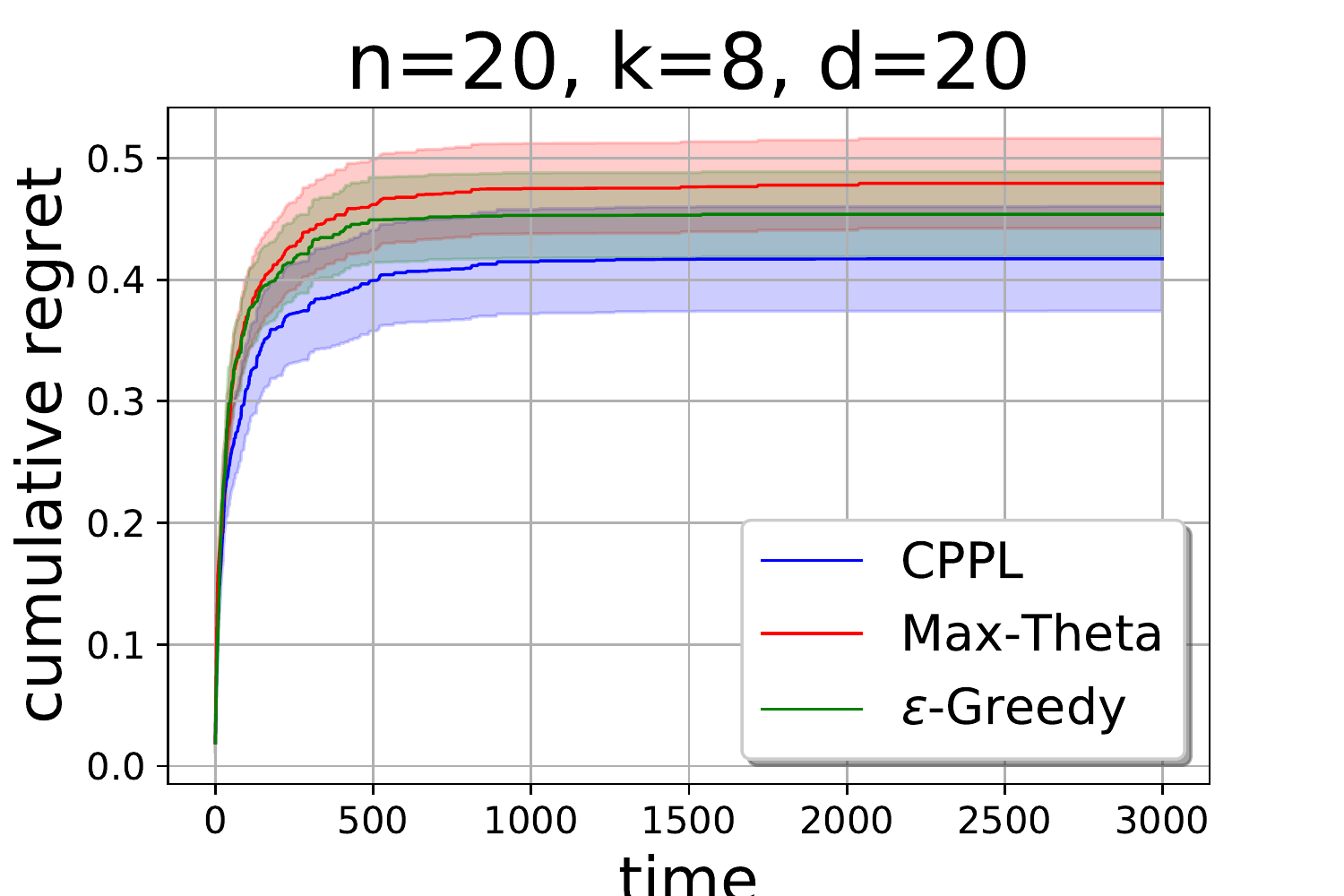}
	\end{subfigure}
	\begin{subfigure}{.333\textwidth}
		\centering
		\includegraphics[width=.99\linewidth]{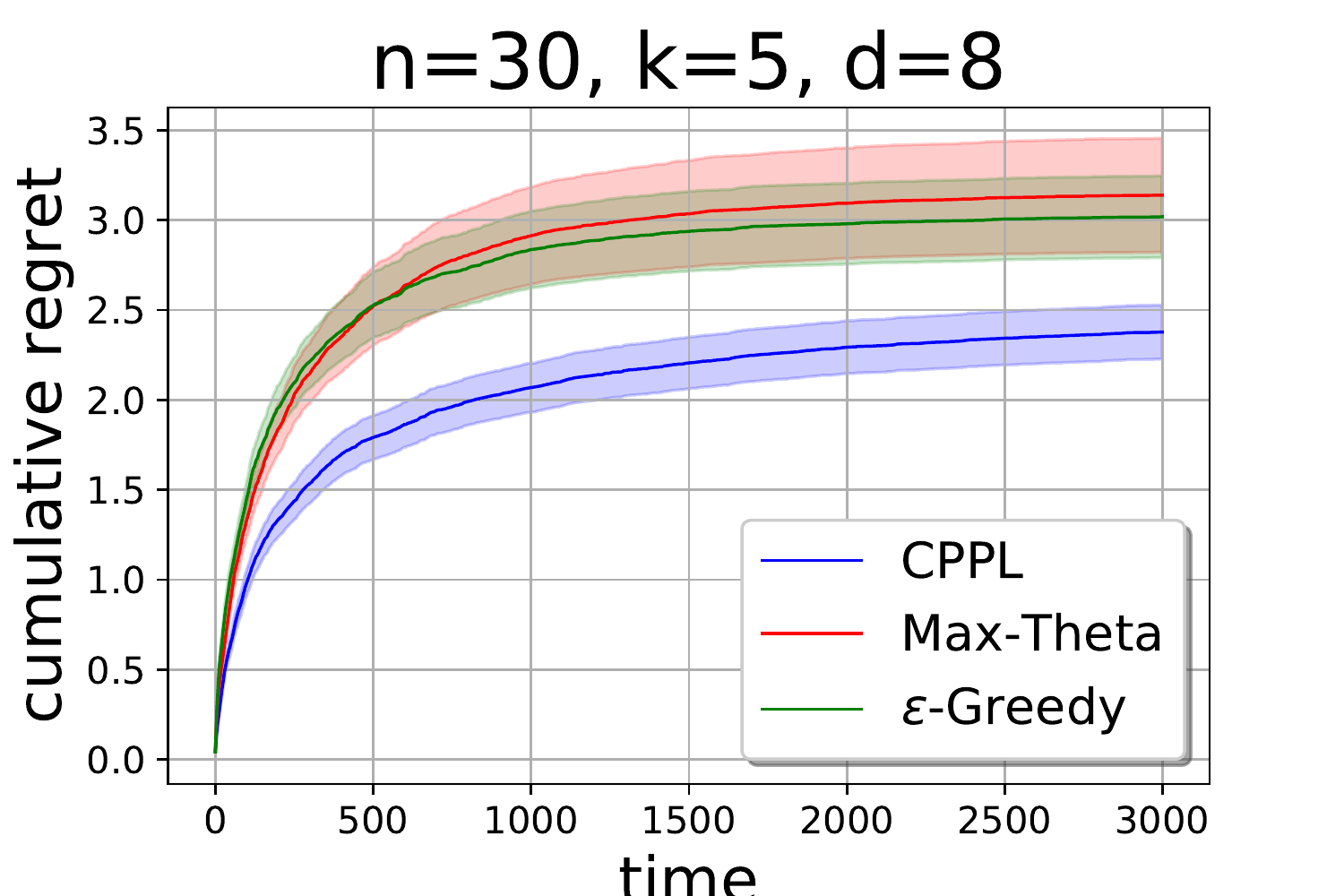}
	\end{subfigure}%
	\begin{subfigure}{.333\textwidth}
		\centering
		\includegraphics[width=.99\linewidth]{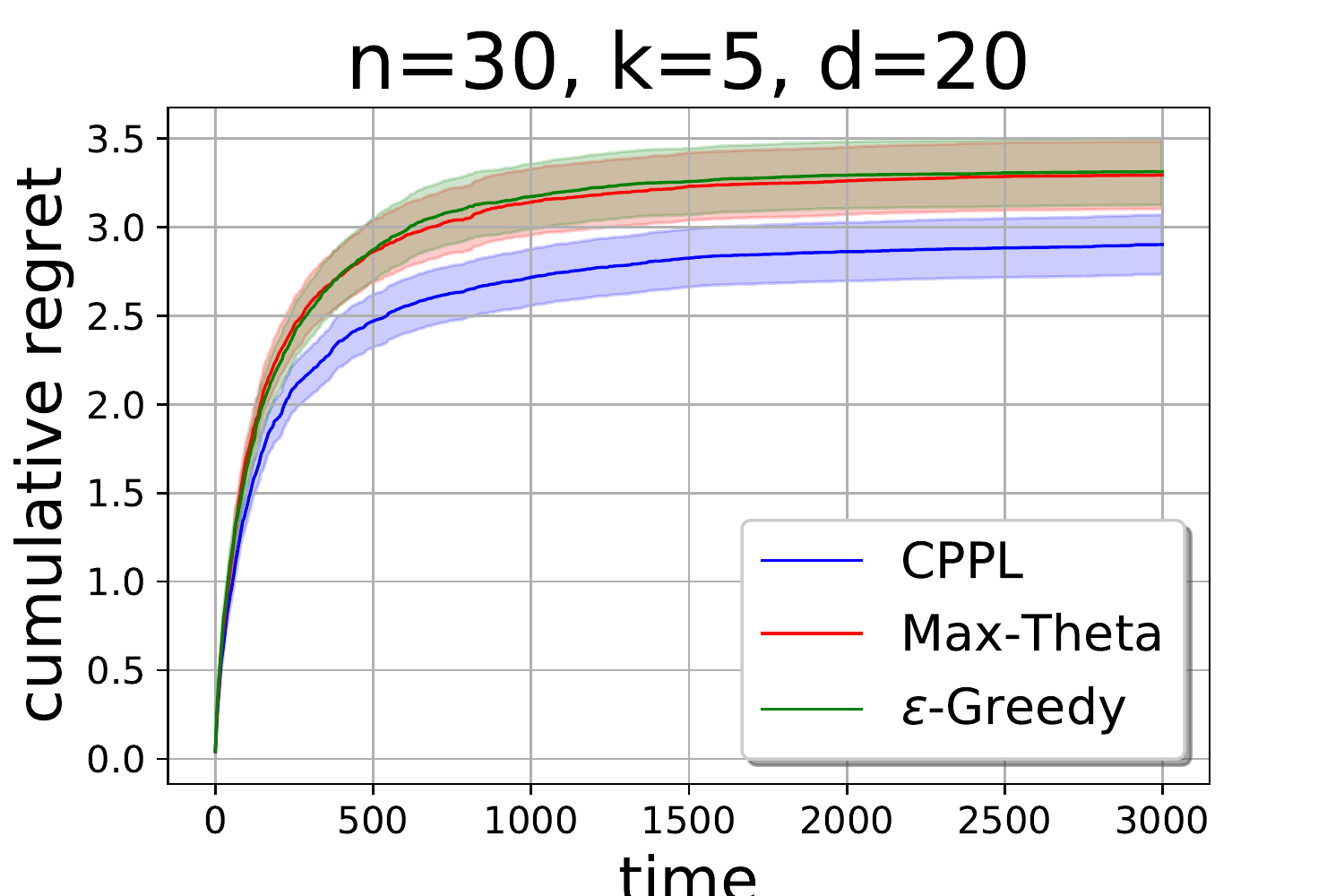}
	\end{subfigure}
	\begin{subfigure}{.333\textwidth}
		\centering
		\includegraphics[width=.99\linewidth]{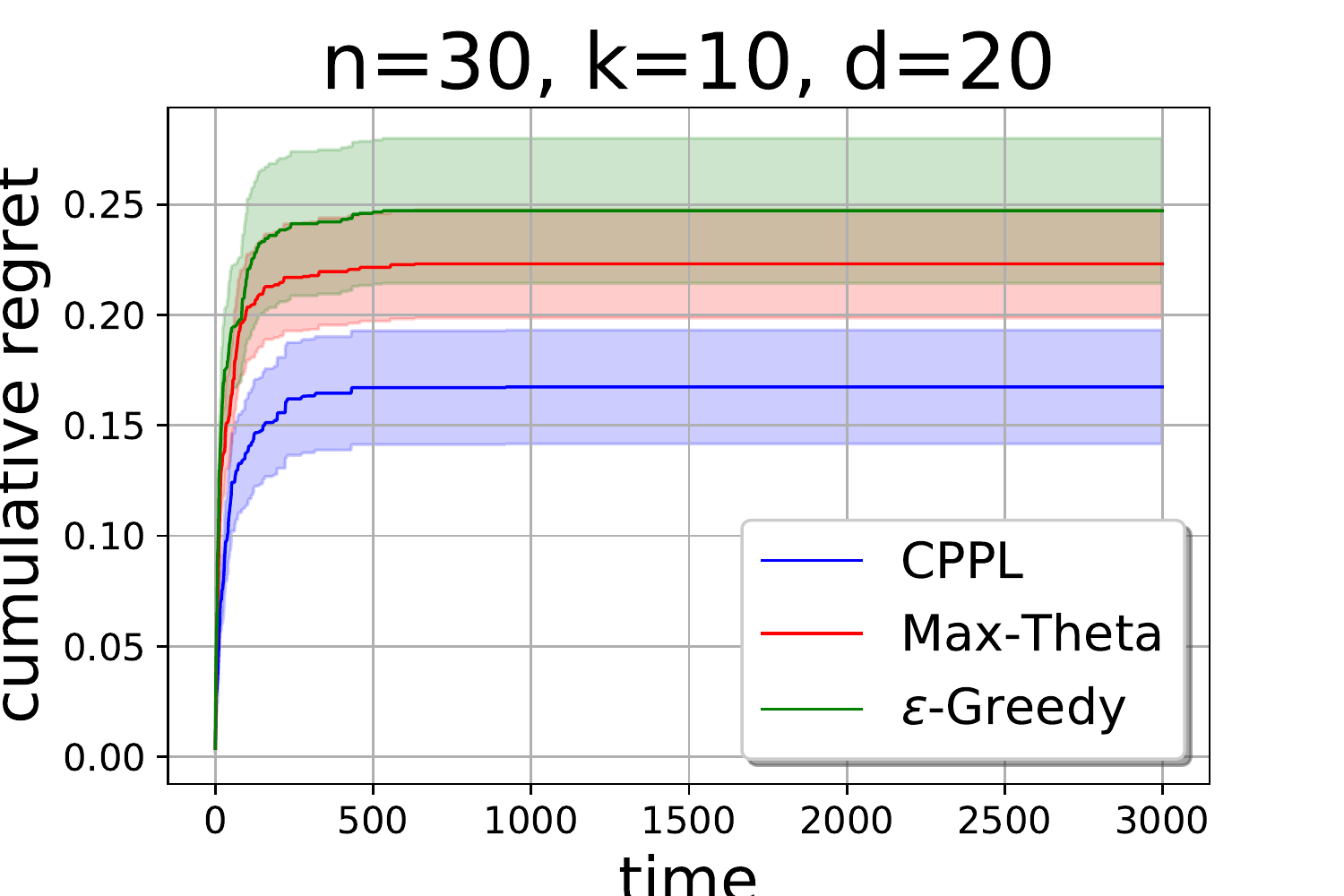}
	\end{subfigure}
	\caption{Cumulative regret of different methods for different parameter settings. The results are averaged over $100$ random repetitions.}
	\label{expSynth}
\end{figure}

The results are illustrated in Figure \ref{expSynth}, where the cumulative regret incurred by the algorithms averaged over 100 repetitions together with the standard error is plotted.
Note that the context-free \Algo{MM} algorithm is omitted due to linear regret in all cases.
%
%
%
For all scenarios, the \Algo{CPPL} method significantly outperforms the other methods, and the superiority increases with the number of available arms $n.$

All displayed scenarios were run under the winner feedback model scenario. The results for the partial ranking feedback scenario are very similar and therefore omitted. 

\subsection{Algorithm Selection}

We also compared \Algo{CPPL}, \Algo{Max-Theta}, \Algo{$\epsilon$-Greedy}, and the MM method  in an algorithm selection scenario, namely the task to select the most efficient algorithm for solving an instance of the propositional satisfiability (SAT) problem. 

As a pool of candidate algorithms, 20 variants of the SAPS solver \citep{tompkins2004scaling} were produced through randomly chosen parametrizations. 
For the sake of reproducibility, we report these parametrizations in the supplement.
Note that each parametrization corresponds to a choice alternative resp.\ arm in our terminology, while the parametrization itself serves as a feature vector (of length 4) for the arm.

Next, we use the last 5000 problem instances from the sat$\_ $SWGCP folder of the AClib\footnote{\url{http://www.aclib.net}.}. 
Each instance is described by a feature vector of length 28, which was computed using SATzilla\footnote{\url{http://www.cs.ubc.ca/labs/beta/Projects/SATzilla}.}. 
We use the ubcsat framework \citep{ToH05} to compute the required running time $\mathbf R_{s,i}$ for each instance-parametrization-combination $(s,i)$, that is, the time required by the solver with parametrization $i$ to terminate on the problem instance $s.$

Prior to running the algorithms, we employ a preprocessing step on the features of the instances, in which we first normalize the features so that they lie in the range $[0,1]$. Then, we remove features with low variance (lower than $0.01$) and those that are highly correlated with others. The last step is performed iteratively in a greedy manner: In each iteration, one of the two most highly correlated features (among those that still remain) is removed, provided the correlation exceeds a predefined threshold ($0.95$). Finally, we end up with a feature representation of size 7. 

In each time step $t$, we randomly pick a problem instance without replacement from the pool of 5000 available problem instances and generate a partial ranking for the selected arms of the algorithms according to a PL model (\ref{def:marginals_context}) with utility parameters set to be $v_i=\exp(-\lambda \mathbf R_{t,i}),$ where $\lambda=10$ and $\mathbf R_{t,i}$ is the running time for the parametrization $i$ of the solver on the problem instance at time step $t.$
Note that the utility parameters thus defined are inversely related to the running times $\mathbf R_{t,i}$.
%
%

\begin{figure}
	\vspace{\figureBetweenHorizontal}
	\centering
	\begin{subfigure}{.3\textwidth}
		\centering
		\includegraphics[width=.99\linewidth]{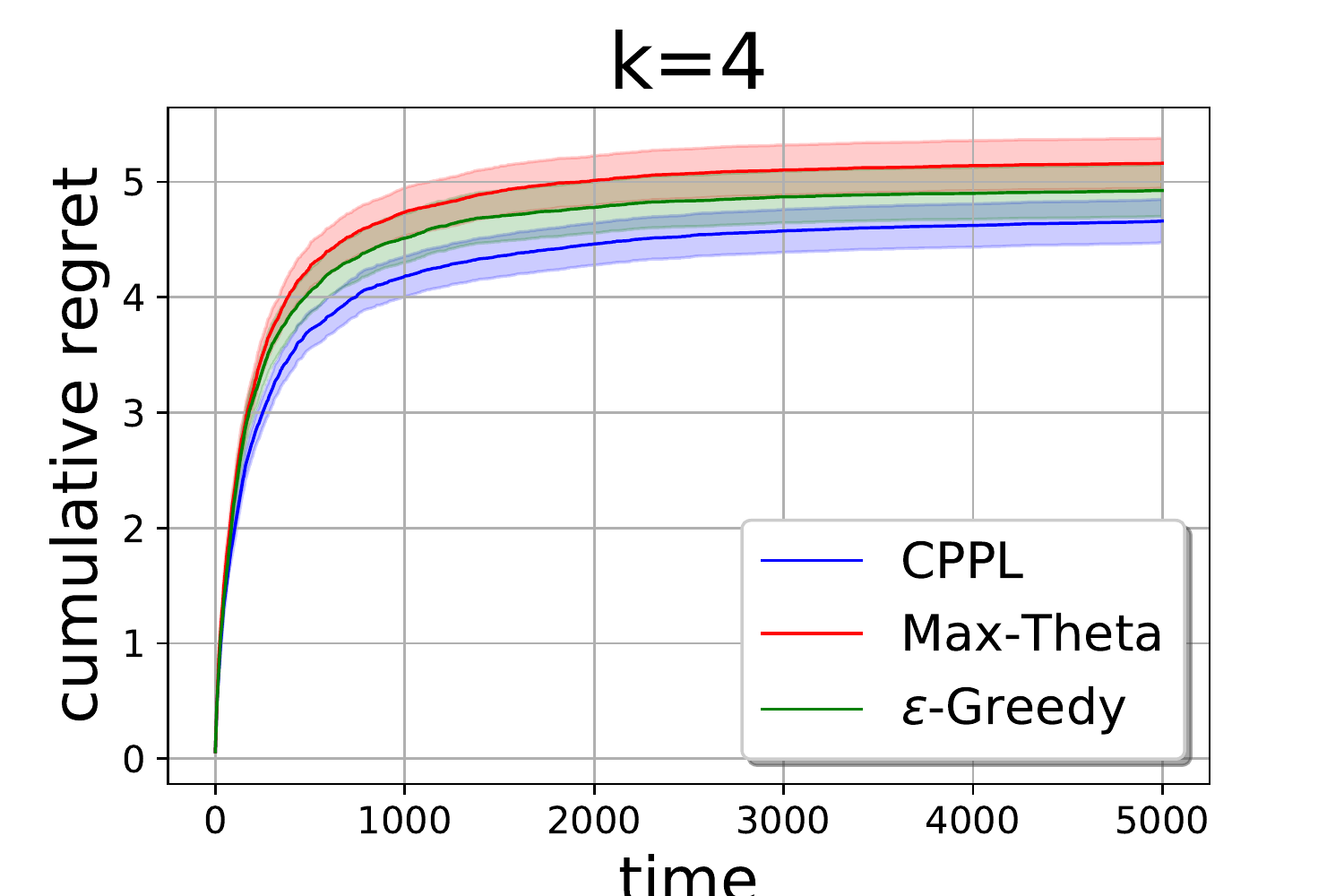}
	\end{subfigure}%
	\begin{subfigure}{.3\textwidth}
		\centering
		\includegraphics[width=.99\linewidth]{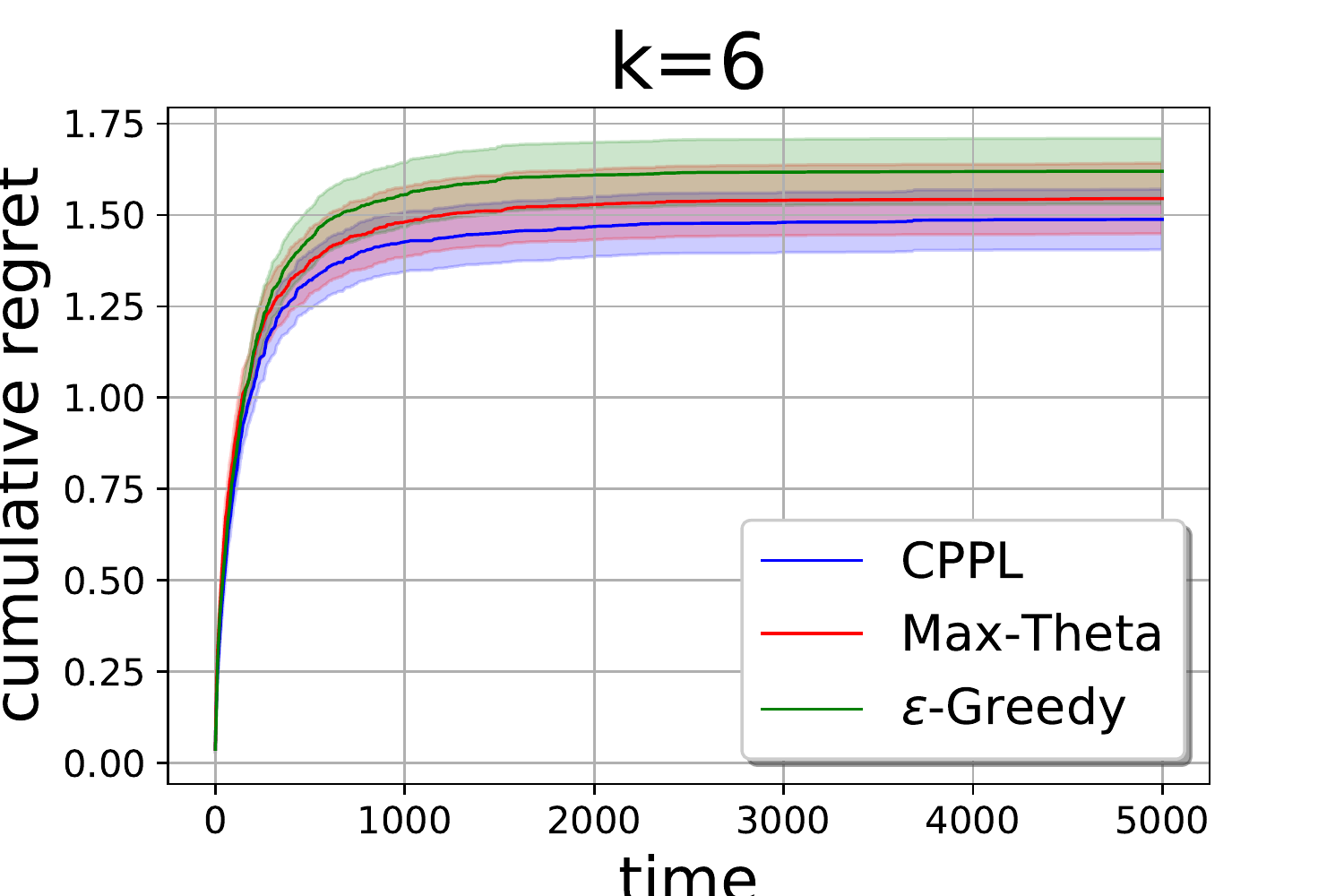}
	\end{subfigure}
	\begin{subfigure}{.3\textwidth}
		\centering
		\includegraphics[width=.99\linewidth]{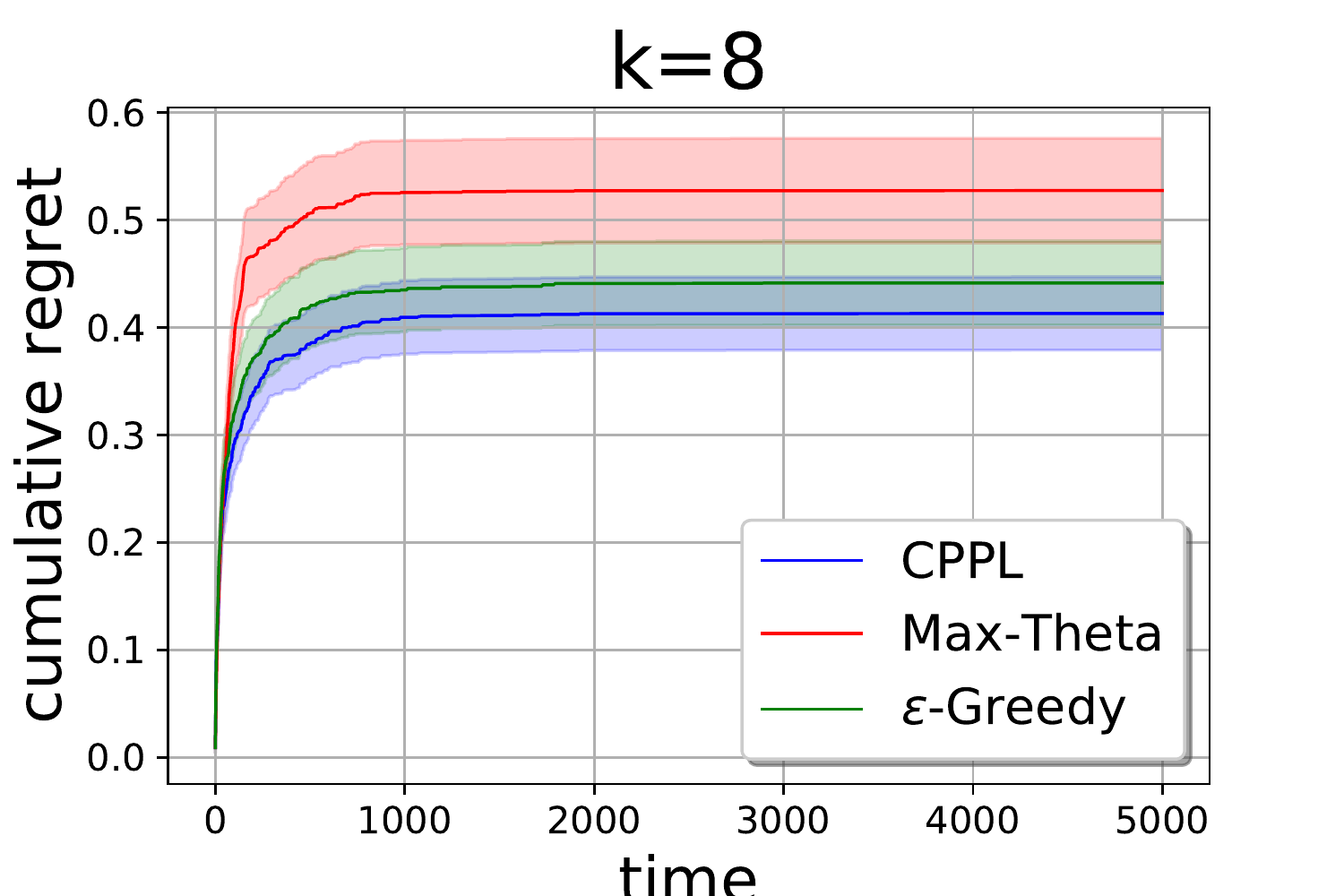}
	\end{subfigure}
	\caption{Cumulative regret of different methods for different $k$'s on an algorithm preselection problem. The results are averaged over $50$ random repetitions.}
	\label{expSaps}
\end{figure}

The result of the experiment is shown in Figure \ref{expSaps}, in which we again used the hyper-parameters $\gamma_1 = 2, \alpha = 0.6, \omega =1$ for our \Algo{CPPL} algorithm (and again omitted the \Algo{MM} algorithm due to linear regret in all cases).
The context vectors were defined in terms of the Kronecker product of the features of the problem instances and the features of the parametrizations.
The findings are similar to those for the experiments on synthetic data, that is, our algorithm outperforms the baselines. 

\section{Conclusion and Future Work}        
We introduced the setting of online preselection under the contextual PL model and proposed an algorithm inspired by the UCB method for effective learning in this setting. The superiority of our algorithm over other online selection methods was demonstrated in experiments on synthetic data and an instance of the algorithm selection problem.

For future work, we are planning to provide a sound theoretical analysis of the considered problem with a sharp lower bound on the regret. Moreover, it would be tempting to show an upper bound on the regret of our proposed algorithm, which matches the lower bound up to constant factors. As explained above, these two problems are very challenging, but we believe there is a realistic prospect of success. 

Moreover, we plan to test our approach on a broader spectrum of algorithm selection problems. Apart form that, our approach can be extended toward handling the problem of online top-$k$ set selection, where the goal is not only to find the set that includes the best arm, but the set of top-$k$ arms. Corresponding algorithms could be applied in information retrieval tasks, such as finding the document most relevant to a given query, or the image most similar to a given query image.



\medskip
\bibliographystyle{chicago}
\bibliography{preselection}

\begin{thebibliography}{}

\bibitem[\protect\citeauthoryear{Abbasi-Yadkori, P\'{a}l, and
  Szepesv\'{a}ri}{Abbasi-Yadkori et~al.}{2011}]{AbPaSz11}
Abbasi-Yadkori, Y., D.~P\'{a}l, and C.~Szepesv\'{a}ri (2011).
\newblock {Improved Algorithms for Linear Stochastic Bandits}.
\newblock In {\em NIPS}, pp.\  2312--2320.

\bibitem[\protect\citeauthoryear{Agarwal, Hsu, Kale, Langford, Li, and
  Schapire}{Agarwal et~al.}{2014}]{AgHsKaLaLiSc14}
Agarwal, A., D.~Hsu, S.~Kale, J.~Langford, L.~Li, and R.~E. Schapire (2014).
\newblock {Taming the Monster: A Fast and Simple Algorithm for Contextual
  Bandits}.
\newblock In {\em ICML}, pp.\  II--1638--II--1646.

\bibitem[\protect\citeauthoryear{Agrawal and Goyal}{Agrawal and
  Goyal}{2013}]{AgGo13}
Agrawal, S. and N.~Goyal (2013).
\newblock {Thompson Sampling for Contextual Bandits with Linear Payoffs}.
\newblock In {\em ICML}, pp.\  III--1220--III--1228.

\bibitem[\protect\citeauthoryear{Auer}{Auer}{2002}]{Au02}
Auer, P. (2002).
\newblock {Using Confidence Bounds for Exploitation-exploration Trade-offs}.
\newblock {\em Journal of Machine Learning Research\/}~{\em 3}, 397--422.

\bibitem[\protect\citeauthoryear{Bengs and H{\"u}llermeier}{Bengs and
  H{\"u}llermeier}{2019}]{bengs19}
Bengs, V. and E.~H{\"u}llermeier (2019).
\newblock Preselection bandits under the plackett-luce model.
\newblock {\em arXiv preprint arXiv:1907.06123\/}.

\bibitem[\protect\citeauthoryear{Busa-Fekete, H\"{u}llermeier, and
  El~Mesaoudi-Paul}{Busa-Fekete et~al.}{2018}]{BuHuEl18}
Busa-Fekete, R., E.~H\"{u}llermeier, and A.~El~Mesaoudi-Paul (2018).
\newblock {Preference-based Online Learning with Dueling Bandits: A survey}.
\newblock {\em CoRR\/}~{\em abs/1807.11398}.

\bibitem[\protect\citeauthoryear{Caro and Gallien}{Caro and
  Gallien}{2007}]{caro2007dynamic}
Caro, F. and J.~Gallien (2007).
\newblock Dynamic assortment with demand learning for seasonal consumer goods.
\newblock {\em Management Science\/}~{\em 53\/}(2), 276--292.

\bibitem[\protect\citeauthoryear{Chen and Frazier}{Chen and
  Frazier}{2017}]{chen2017dueling}
Chen, B. and P.~I. Frazier (2017).
\newblock Dueling bandits with weak regret.
\newblock In {\em ICML}, pp.\  731--739.

\bibitem[\protect\citeauthoryear{Chen, Li, and Mao}{Chen
  et~al.}{2018}]{ChLiMa18}
Chen, X., Y.~Li, and J.~Mao (2018).
\newblock {A Nearly Instance Optimal Algorithm for Top-k Ranking Under the
  Multinomial Logit Model}.
\newblock In {\em Annual {ACM-SIAM} Symposium on Discrete Algorithms {(SODA)}},
  pp.\  2504--2522.

\bibitem[\protect\citeauthoryear{Chen, Wang, and Zhou}{Chen
  et~al.}{2018}]{ChWaZh18}
Chen, X., Y.~Wang, and Y.~Zhou (2018).
\newblock {Dynamic Assortment Optimization with Changing Contextual
  Information}.
\newblock {\em CoRR\/}~{\em abs/1810.13069}.

\bibitem[\protect\citeauthoryear{Cheng, H{\"u}llermeier, and Dembczynski}{Cheng
  et~al.}{2010}]{cheng2010label}
Cheng, W., E.~H{\"u}llermeier, and K.~J. Dembczynski (2010).
\newblock Label ranking methods based on the plackett-luce model.
\newblock In {\em ICML}, pp.\  215--222.

\bibitem[\protect\citeauthoryear{Chi~Cheung and Simchi-levi}{Chi~Cheung and
  Simchi-levi}{2017}]{ChSi17}
Chi~Cheung, W. and D.~Simchi-levi (2017).
\newblock {Thompson Sampling for Online Personalized Assortment Optimization
  Problems with Multinomial Logit Choice Models}.
\newblock {\em SSRN Electronic Journal\/}.

\bibitem[\protect\citeauthoryear{Chu, Li, Reyzin, and Schapire}{Chu
  et~al.}{2011}]{ChLiReSc11}
Chu, W., L.~Li, L.~Reyzin, and R.~E. Schapire (2011).
\newblock {Contextual Bandits with Linear Payoff Functions}.
\newblock In {\em {(AISTATS)}}, pp.\  208--214.

\bibitem[\protect\citeauthoryear{Cohen and Crammer}{Cohen and
  Crammer}{2014}]{CoCr14}
Cohen, H. and K.~Crammer (2014).
\newblock {Learning Multiple Tasks in Parallel with a Shared Annotator}.
\newblock In {\em Advances in Neural Information Processing Systems {(NIPS)}},
  pp.\  1170--1178.

\bibitem[\protect\citeauthoryear{Dud{\'\i}k, Hofmann, Schapire, Slivkins, and
  Zoghi}{Dud{\'\i}k et~al.}{2015}]{DuHoShSlZo15}
Dud{\'\i}k, M., K.~Hofmann, R.~E. Schapire, A.~Slivkins, and M.~Zoghi (2015).
\newblock {Contextual Dueling Bandits}.
\newblock In {\em COLT}, pp.\  563--587.

\bibitem[\protect\citeauthoryear{Fang, Xu, and Yang}{Fang
  et~al.}{2018}]{FaXuYa18}
Fang, Y., J.~Xu, and L.~Yang (2018).
\newblock {Online Bootstrap Confidence Intervals for the Stochastic Gradient
  Descent Estimator}.
\newblock {\em Journal of Machine Learning Research\/}~{\em 19\/}(78), 1--21.

\bibitem[\protect\citeauthoryear{Hunter}{Hunter}{2004}]{hunter2004mm}
Hunter, D.~R. (2004).
\newblock {MM Algorithms for Generalized Bradley-Terry Models}.
\newblock {\em The Annals of Statistics\/}~{\em 32\/}(1), 384--406.

\bibitem[\protect\citeauthoryear{Johnstone}{Johnstone}{2001}]{johnstone2001chi}
Johnstone, I.~M. (2001).
\newblock Chi-square oracle inequalities.
\newblock {\em Lecture Notes-Monograph Series\/}, 399--418.

\bibitem[\protect\citeauthoryear{Kerschke, Hoos, Neumann, and
  Trautmann}{Kerschke et~al.}{2018}]{Kerschke18}
Kerschke, P., H.~H. Hoos, F.~Neumann, and H.~Trautmann (2018).
\newblock Automated algorithm selection: {S}urvey and perspectives.
\newblock {\em arXiv preprint arXiv:1811.11597\/}.

\bibitem[\protect\citeauthoryear{Langford and Zhang}{Langford and
  Zhang}{2007}]{La07}
Langford, J. and T.~Zhang (2007).
\newblock {The Epoch-Greedy Algorithm for Contextual Multi-armed Bandits}.
\newblock In {\em NIPS}, pp.\  1--8.

\bibitem[\protect\citeauthoryear{Lattimore and Szepesv{\'a}ri}{Lattimore and
  Szepesv{\'a}ri}{2019}]{LaSz18}
Lattimore, T. and C.~Szepesv{\'a}ri (2019).
\newblock {Bandit Algorithms}.
\newblock {\em Preprint\/}.

\bibitem[\protect\citeauthoryear{Laurent and Massart}{Laurent and
  Massart}{2000}]{LaMa00}
Laurent, B. and P.~Massart (2000).
\newblock {Adaptive Estimation of a Quadratic Functional by Model Selection}.
\newblock {\em Annals of Statistics\/}~{\em 28\/}(5), 1302--1338.

\bibitem[\protect\citeauthoryear{Li, Chu, Langford, and Schapire}{Li
  et~al.}{2010}]{LiChLaSc10}
Li, L., W.~Chu, J.~Langford, and R.~E. Schapire (2010).
\newblock {A Contextual-Bandit Approach to Personalized News Article
  Recommendation}.
\newblock In {\em International Conference on World Wide Web}, pp.\  661--670.
  ACM.

\bibitem[\protect\citeauthoryear{Lu, Pal, and Pal}{Lu et~al.}{2010}]{LuPaPa10}
Lu, T., D.~Pal, and M.~Pal (2010).
\newblock {Contextual Multi-Armed Bandits}.
\newblock In {\em International Conference on Artificial Intelligence and
  Statistics {(AISTATS)}}, pp.\  485--492.

\bibitem[\protect\citeauthoryear{Ou, Li, Zhu, and Jin}{Ou
  et~al.}{2018}]{OuLiZhJi18}
Ou, M., N.~Li, S.~Zhu, and R.~Jin (2018).
\newblock {Multinomial Logit Bandit with Linear Utility Functions}.
\newblock In {\em IJCAI}, pp.\  2602--2608.

\bibitem[\protect\citeauthoryear{Polyak and Juditsky}{Polyak and
  Juditsky}{1992}]{PoJu92}
Polyak, B. and A.~Juditsky (1992).
\newblock {Acceleration of Stochastic Approximation by Averaging}.
\newblock {\em SIAM J. Control Optim.\/}~{\em 30\/}(4), 838--855.

\bibitem[\protect\citeauthoryear{Ren, Liu, and Shroff}{Ren
  et~al.}{2018}]{ReLiSh18}
Ren, W., J.~Liu, and N.~B. Shroff (2018).
\newblock {PAC Ranking from Pairwise and Listwise Queries: Lower Bounds and
  Upper Bounds}.
\newblock {\em CoRR\/}~{\em abs/1806.02970}.

\bibitem[\protect\citeauthoryear{Ruppert}{Ruppert}{1988}]{Ru88}
Ruppert, D. (1988).
\newblock {Efficient Estimations from a Slowly Convergent Robbins-Monro
  Process}.
\newblock Technical report, {Cornell University Operations Research and
  Industrial Engineering}.

\bibitem[\protect\citeauthoryear{Saha and Gopalan}{Saha and
  Gopalan}{2018a}]{SaGo18b}
Saha, A. and A.~Gopalan (2018a).
\newblock {Active Ranking with Subset-wise Preferences}.
\newblock {\em CoRR\/}~{\em abs/1810.10321}.

\bibitem[\protect\citeauthoryear{Saha and Gopalan}{Saha and
  Gopalan}{2018b}]{SaGo18a}
Saha, A. and A.~Gopalan (2018b).
\newblock {Battle of Bandits}.
\newblock In {\em UAI}.

\bibitem[\protect\citeauthoryear{Saha and Gopalan}{Saha and
  Gopalan}{2019a}]{SaGo19a}
Saha, A. and A.~Gopalan (2019a).
\newblock {PAC Battling Bandits in the Plackett-Luce Model}.
\newblock In {\em International Conference on Algorithmic Learning Theory
  {(ALT)}}, pp.\  700--737.

\bibitem[\protect\citeauthoryear{Saha and Gopalan}{Saha and
  Gopalan}{2019b}]{SaGo19b}
Saha, A. and A.~Gopalan (2019b).
\newblock {Regret Minimisation in Multinomial Logit Bandits}.
\newblock {\em CoRR\/}~{\em abs/1903.00543}.

\bibitem[\protect\citeauthoryear{Sch{\"a}fer and H{\"u}llermeier}{Sch{\"a}fer
  and H{\"u}llermeier}{2018}]{ScHu18}
Sch{\"a}fer, D. and E.~H{\"u}llermeier (2018).
\newblock {Dyad Ranking Using Plackett--Luce Models Based on Joint Feature
  Representations}.
\newblock {\em Machine Learning\/}~{\em 107\/}(5), 903--941.

\bibitem[\protect\citeauthoryear{Tompkins and Hoos}{Tompkins and
  Hoos}{2005}]{ToH05}
Tompkins, D. A.~D. and H.~H. Hoos (2005).
\newblock Ubcsat: An implementation and experimentation environment for sls
  algorithms for sat and max-sat.
\newblock In {\em {Revised Selected Papers from ({SAT 2004})}}, Volume 3542,
  pp.\  306--320.

\bibitem[\protect\citeauthoryear{Tompkins, Hutter, and Hoos}{Tompkins
  et~al.}{2004}]{tompkins2004scaling}
Tompkins, D. A.~D., F.~Hutter, and H.~H. Hoos (2004).
\newblock Scaling and probabilistic smoothing (saps).
\newblock {\em SAT\/}.

\bibitem[\protect\citeauthoryear{Yue and Joachims}{Yue and
  Joachims}{2009}]{YuJo09}
Yue, Y. and T.~Joachims (2009).
\newblock {Interactively Optimizing Information Retrieval Systems As a Dueling
  Bandits Problem}.
\newblock In {\em {(ICML)}}, pp.\  1201--1208.

\end{thebibliography}

\newpage
\appendix

\begin{center}
	{\LARGE Supplementary material}
\end{center}

\section{Gradient and Hessian Matrix of the Log-Likelihood function}

The gradient of the log-likelihood function in the partial ranking feedback scenario (\ref{def:log_likelihood}) is given by
\begin{flalign*} 
\begin{split}
&\nabla~\cL \left( \theta \vert \sigma, S, \bX\right) = \sum_{i=1}^{|S|} \left[ \bx_{\sigma^{-1}(i)} -  \frac{a_i(\theta|\sigma,S,\bX)  }{b_i(\theta|\sigma,S,\bX)} \right] \enspace, \\
%
\end{split}
\end{flalign*}  
and its Hessian Matrix is 
\begin{align*} 
\begin{split}
\nabla^2 ~\cL &\left( \theta \vert \sigma, S, \bX\right) 
 = \sum_{i=1}^{|S|} \left[	\frac{ a_i(\theta|\sigma,S,\bX) (a_i(\theta|\sigma,S,\bX))^{T} }{ (	b_i(\theta|\sigma,S,\bX))^{2} }	 \right]  
 - \sum_{i=1}^{|S|} \left[	\frac{ c_i(\theta|\sigma,S,\bX) }{	b_i(\theta|\sigma,S,\bX) } \right],
\end{split}
\end{align*}
where for each $i\in \{1,\ldots,|S|\}$ we defined
\begin{align*}
%
a_i(\theta|\sigma,S,\bX) &= \sum_{l=i}^{|S|} \bx_{\sigma^{-1}(l)}  \exp(\theta^{\top} \bx_{\sigma^{-1}(l)}  ), \\
b_i(\theta|\sigma,S,\bX) &= \sum_{l=i}^{|S|} \exp(\theta^{\top} \bx_{\sigma^{-1}(l)} ), \\
c_i(\theta|\sigma,S,\bX)  &= \sum_{l=i}^{|S|}\exp(\theta^{\top} \bx_{\sigma^{-1}(l)}  )  \bx_{\sigma^{-1}(l)} \bx_{\sigma^{-1}(l)}^{\top}.
\end{align*}
The gradient of the log-likelihood function in the partial winner feedback scenario (\ref{def:log_likelihood_top_rank}) is given by
\begin{align*}
&\qquad \nabla~\cL \left( \theta \vert k, S, \bX\right) = \bx_{k} -  \frac{a(\theta|S,\bX) }{  b(\theta|S,\bX) }  \enspace, 
\end{align*}
and its Hessian matrix is
\begin{align*}
\nabla^2 ~\cL \left( \theta \vert k, S, \bX\right) & = 
\left[ \frac{ a(\theta|S,\bX) (a(\theta|S,\bX))^{T} }{  (b(\theta|S,\bX))^{2}  } \right] 
 - \left[ 	\frac{  c(\theta|S,\bX)  }{ b(\theta|S,\bX)  }	\right],
\end{align*}
where we abbreviated
\begin{align*}
%
a(\theta|S,\bX) &= \sum_{l \in S} \bx_{l}  \exp(\theta^{\top} \bx_{l}  ), \\
b(\theta|S,\bX)&= \sum_{l \in S} \exp(\theta^{\top} \bx_{l}  ), \\
c(\theta|S,\bX) &= \sum_{l \in S}\exp(\theta^{\top} \bx_{l}  )  \bx_{l} \bx_{l}^{\top} \enspace .
\end{align*}

\section{Proof of Lemma \ref{eq:F_distr_tails}}

Note that an F-distributed random variable $X$ with degrees of freedom $d_1$ and $d_2$ has the same distribution as the ratio between $Y_1/d_1$ and $Y_2/d_2,$ where $Y_1,Y_2$ are independent random variables with $\chi^2$-distributions and degrees of freedom $d_1$ resp.\ $d_2.$
For a random variable $Y$ with a $\chi^2$-distribution with $d$ degrees of freedom, the following tail bounds hold \citep{LaMa00,johnstone2001chi}:
\begin{align} \label{eq:chi_square_tails}
\begin{split}
&\prob( Y - d \geq 2\sqrt{dx} + 2x) \leq \exp(-x) \, , \qquad \forall x\geq0 \, , \\
&\prob( |Y - d| \geq d\, x  ) \leq \exp(-\nicefrac{3\, d\, x^2 }{16}) \, , \qquad x\in[0,\nicefrac12) \, .
\end{split}
\end{align}
For sake of brevity write $s=\frac{ 4(d_1 + 2\sqrt{d_1\,x} + 2x)}{3\, d_1}.$
Then, using these tail bounds, one can derive that
\begin{flalign*}
&\prob\left( X \geq s \right) \leq 	\prob\left( Y_1 \geq \frac34 d_1 s \right) + \prob\left( \frac{1}{Y_2} \geq  \frac{4}{3\, d_2}  \right) 
=: (i) + (ii) \enspace .
\end{flalign*} 
%
By the first inequality in (\ref{eq:chi_square_tails}), the first term $(i)$ can be bounded  by $\exp(-x),$ while the second term $(ii)$ can be bounded as follows:
\begin{align*}
(2) &= \prob\left( \frac{1}{4} \leq 1-  \frac{Y_2}{d_2}  \right) \leq \exp\left( - \frac{3\, d_2}{2^8} \right),
\end{align*}
where we used the second inequality in (\ref{eq:chi_square_tails}) for deriving the last inequality in the latter display.

\begin{table}
	\caption{The parametrizations used for the SAT solver.}
	\label{table_params}
	\begin{center}
		\resizebox{0.4\textwidth}{!}{
			\begin{tabular}{||c | c | c | c||} 
				\hline
				$\alpha$ & $\rho$ & $ps$ & $wp$ \\ [0.5ex] 
				\hline\hline
				1.54114	& 0.851212 & 0.739441 & 0.846641 \\
				\hline
				1.85872	& 0.662701	& 0.532759	& 0.193693 \\
				\hline
				1.48834	& 0.839351	& -0.219562	& 0.973629 \\
				\hline
				0.807838 &	0.787876	& 0.634953	& 0.987749 \\
				\hline
				1.27937	& 0.78502	& 0.871624	& 0.769004 \\
				\hline
				1.08792	& 0.0999449	& 0.624429	& -0.271802 \\
				\hline
				1.65351	& 0.828302	& 0.711434	& 0.952419 \\
				\hline
				1.45122	& 0.515806	& 0.804639	& -0.202731 \\
				\hline
				1.6245	& 0.553748	& 0.464212	& 0.419496 \\
				\hline
				1.40361	& 0.728122	& 0.743332	& 0.733345 \\
				\hline
				1.92961	& 0.91072	& 0.995255	& 0.958284 \\
				\hline
				1.40955	& 0.740847	& -0.0788468	& -0.367871 \\
				\hline
				1.79481	& 0.82973	& 0.81912	& 0.318974 \\
				\hline
				1.8575	& 0.654704	& 0.588427	& 0.466252 \\
				\hline
				1.74699	& 0.699139	& 0.610026	& 0.96016 \\
				\hline
				1.88793	& 0.464533	& 0.924795	& 0.727195 \\
				\hline
				1.6724	& 0.830823	& 0.704286	& 0.791863 \\
				\hline
				1.97717	& 0.941355	& 0.824751	& 0.935904 \\
				\hline
				1.71295	& -0.955414	& 0.988518	& 0.585664 \\
				\hline
				0.859881 &	0.932354 &	-0.0785504	& 0.496108 \\
				\hline
			\end{tabular}
		}
	\end{center}
\end{table}

\section{Experimental Details}

In Table \ref{table_params} we report the parametrizations of the 20 SAT solver used in Section \ref{sec_exp} for the application of our method for the algorithm selection problem.

\end{document}